\documentclass[10pt,twocolumn,letterpaper]{article}

\usepackage{cvpr}              % To produce the CAMERA-READY version
\usepackage[dvipsnames]{xcolor}
\usepackage{tcolorbox}
\tcbuselibrary{most}
\newtcolorbox{prompt}[1]{
    enhanced,
    drop shadow=black!5!white,
    left=4mm,
    right=4mm,
    top=3mm,
    bottom=3mm,
    boxsep=0mm,
    rounded corners,
    title=#1,
    fontupper=\linespread{1.1}\scriptsize\fontfamily{lmr}\selectfont,
    breakable
}

\newtcolorbox{example}{
    colback=yellow!10,
    colframe=yellow!40,
    fonttitle=\bfseries,
    coltitle=black,
    colbacktitle=yellow!60,
    enhanced,
    drop shadow=black!5!white,
    left=2mm,
    right=2mm,
    top=3mm,
    bottom=3mm,
    boxsep=0mm,
    sharp corners,
    float
    }

\definecolor{cvprblue}{rgb}{0.21,0.49,0.74}

\usepackage{graphicx}
\usepackage{float}
\usepackage{booktabs}
\usepackage{multirow}
\usepackage{xcolor,colortbl}
\usepackage{tabularx}
\usepackage{url}

\usepackage[pagebackref,breaklinks,colorlinks,citecolor=cvprblue]{hyperref}

\def\paperID{6624} % *** Enter the Paper ID here
\def\confName{CVPR}
\def\confYear{2025}

\title{VisionArena: 230K Real World User-VLM Conversations with Preference Labels}

\author{%
  \begin{tabular}{cccc}
    Christopher Chou* & Lisa Dunlap* & Koki Mashita & Krishna Mandal \\
    Stanford         & UC Berkeley  & UC Berkeley  & UC Berkeley   \\
    \\
    Trevor Darrell   & Ion Stoica   & Joseph E. Gonzalez & Wei-Lin Chiang \\
    UC Berkeley      & UC Berkeley  & UC Berkeley        & UC Berkeley    \\
  \end{tabular}
}

\newcommand{\dataset}[1]{VisionArena}
\definecolor{midgray}{gray}{0.4}

\begin{document}

\twocolumn[{
\renewcommand\twocolumn[1][]{#1}
\maketitle
\begin{center}
    \vspace{-1em}
    \centering
    \captionsetup{type=figure} 
    \includegraphics[width=\textwidth, trim=0 0 0 0, clip]{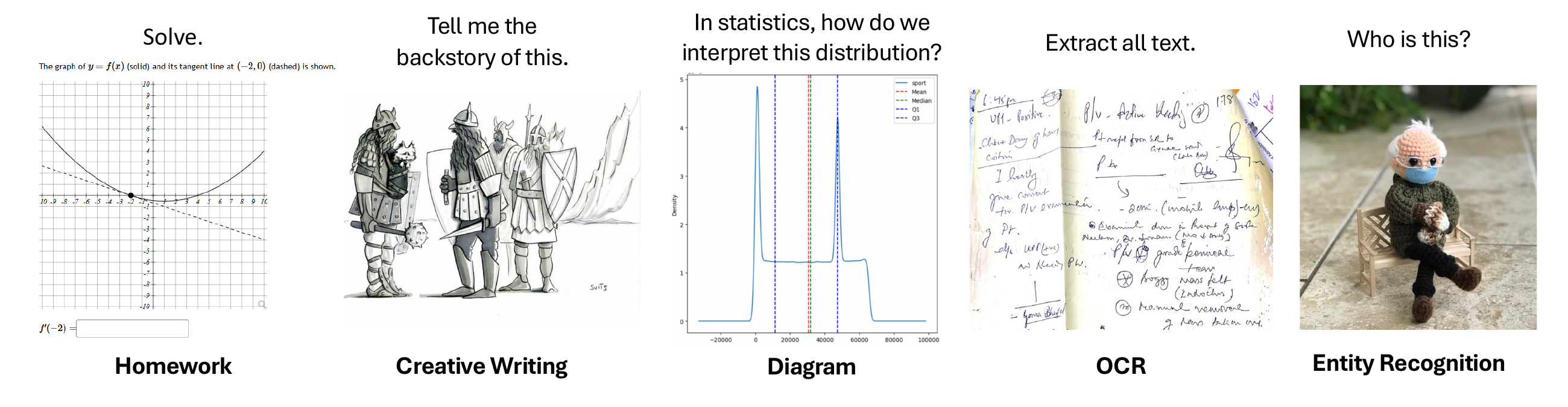}
    \vspace{-2.5em}
    \captionof{figure}{\textbf{Samples from \dataset{} Conversations.}
    \dataset{} contains conversations from real users covering a variety of domains. }
    \label{fig:teaser}
\end{center}
}]

{
  \renewcommand{\thefootnote}
    {\fnsymbol{footnote}}
  \footnotetext[1]{Equal contribution. }
}

\maketitle
\begin{abstract}
With the growing adoption and capabilities of vision-language models (VLMs) comes the need for benchmarks that capture authentic user-VLM interactions. In response, we create \dataset{}, a dataset of 230K real-world conversations between users and VLMs. Collected from Chatbot Arena — an open-source platform where users interact with VLMs and submit preference votes — VisionArena spans 73K unique users, 45 VLMs, and 138 languages. Our dataset contains three subsets: \textbf{VisionArena-Chat}, 200k single and multi-turn conversations between a user and a VLM; \textbf{VisionArena-Battle}, 30K conversations comparing two anonymous VLMs with user preference votes; and \textbf{VisionArena-Bench}, an automatic benchmark of 500 diverse user prompts that efficiently approximate the live Chatbot Arena model rankings. Additionally, we highlight the types of question asked by users, the influence of response style on preference, and areas where models often fail. 
We find open-ended tasks like captioning and humor are highly style-dependent, and current VLMs struggle with spatial reasoning and planning tasks. Lastly, we show finetuning the same base model on VisionArena-Chat outperforms Llava-Instruct-158K, with a 17-point gain on MMMU and a 46-point gain on the WildVision benchmark. Dataset at \url{https://huggingface.co/lmarena-ai}.
% Lastly, we show that running automatic VLM evaluation on VisionArena-Bench generates a model ranking which is consistent with the live Chatbot Arena model rankings.
\end{abstract} 
\vspace{-1em}
\section{Introduction}
\label{sec:intro}

Visual language models (VLMs)~\cite{openai2023gpt4,reid2024gemini,claude32024family,llama3modelcard} are being increasingly used in a wide range of real-world applications including image captioning and story telling, document understanding, web development, and embodied systems. 
While these models have made remarkable progress on a wide range of benchmarks~\cite{VQA,yue2024mmmumassivemultidisciplinemultimodal,balanced_binary_vqa,balanced_vqa_v2,gurari2018vizwiz,gurari2019vizwizpriv},
existing VLM benchmarks focus largely on static, single-turn tasks with predetermined correct answers, overlooking the open-ended, evolving nature of real-world user interactions. They also rarely capture multi-turn dialogue, incorporate diverse context, or reflect the fluidity of user intent. As such, they provide a simplified snapshot of VLM capabilities.

\begin{table*}[h]
\centering
\small
\resizebox{\textwidth}{!}{%
\begin{tabular}{lrrrrrrrrrrr}
\toprule
\multirow{2}{*}{Dataset} & \multirow{2}{*}{\# Convs} & \multirow{2}{*}{\# Models} &  \multirow{2}{*}{\# Users} & \multirow{2}{*}{\# Langs} & \% Unique Images & Avg. \# Turns & Avg. \# Tokens & Avg. \# Tokens & Human   \\
&  &  & &  & per Sample  &  & per Prompt & per Response & Preference\\
\midrule
\midrule
LMSYS-Chat-1M & 1,000,000 & 25 & 13,500 & 35 & - & 2.0 & 36.9 & 214.2 & No & \\
WildVision-Battle & 10,383 & 19 & - & 28 & 56.2 & 1.2 & 57.8 & 131.7 & Yes & \\
WildVision-Chat & 45,170 & 9 & - & 26 & 33.4 & 1.4 & 81.3 & 171.6 & No & \\
\midrule
\dataset{}-Battle & 30,000 & 17 & 14,031 & 90 & 76.4 & 1.3 & 90.2 & 393.6 & Yes & \\
\dataset{}-Chat & 200,000 & 45 & 72.933 & 138 & 62.1 & 1.5 & 184.1 & 634.3 & No & \\
\bottomrule
\end{tabular}%
}
\vspace{-1em}
\caption{\textbf{Dataset Comparison.} Compared to previous VLM preference benchmarks, \dataset{} contains 3x the amount of data, with more users, language, models, unique images, and conversation turns.}
\label{tab:dataset_comparrison}
\end{table*}

Understanding these real-world interactions across a variety of tasks is essential for developing models that align with human expectations and perform effectively.
% across diverse scenarios. 
To address this, previous works such as Chatbot Arena~\cite{chiang2024chatbot} and WildVision~\cite{lu2024wildvision,yujie2024wildvisionv2} crowdsource evaluation by hosting platforms where users can freely interact with pairs of VLMs and provide preference votes. 
Building off of these works, we introduce \textbf{\dataset{}}, a dataset of 230K real-world conversations between users and 38 VLMs in 135 languages, collected through the Chatbot Arena platform. \dataset{} consists of:

\begin{itemize}
    \item \textbf{\dataset{}-Chat}: 200,000 single and multi-turn chat logs between users and VLMs, spanning 138 languages, 73k users, and 45 open source and proprietary VLMs.
    \item \textbf{\dataset{}-Battle}: 30,000 conversations where users interact with two anonymized VLMs, along with preference votes indicating which response they prefer.
    \item \textbf{\dataset{}-Bench}: An automatic benchmark consisting of 500 diverse user prompts that can be used to cheaply approximate model rankings via automatic benchmarking with VLM as a judge. 
\end{itemize}

We conduct analysis of these datasets and construct a set of popular question categories including captioning, OCR, humor, creative writing, entity recognition, and diagram understanding. We also explore the influence of stylistic properties of responses such as response length, markdown, and specificity on human preference. We find that more open-ended questions like captioning and humor are heavily influenced by style, which causes certain models like InternVL to have a disproportionately higher ranking in these categories. 
We provide this metadata with \dataset{} to enable further analysis. We also highlight common failure modes of VLMs and provide a small curated set of user prompts where top proprietary models fail, including complex spatial reasoning and planning tasks.

Next, we demonstrate how \dataset{} can be used to improve VLMs through instruction finetuning. Compared to LLaVA-Instruct-158K~\cite{liu2023visualinstructiontuning}, by finetuning on data from VisionArena-Chat, models show a 17 point improvement in MMMU~\cite{yue2024mmmumassivemultidisciplinemultimodal} and a 46 point improvement on the human preference benchmark WV-Bench~\cite{lu2024wildvision}. In addition to \dataset{}, we also release this finetuned model. 
Lastly, we build on existing work in automatic benchmarking of VLMs to show that evaluating on the 500 prompts in VisionArena-Bench results in a model lineup that is consistent with the much larger online preference leaderboard Chatbot Arena. When compared with other automatic preference benchmarks like WildVision-Bench,  \dataset{}-Bench is far more predictive of the online Chatbot Arena VLM leaderboard performance, which contains over 100,000 user votes as of October 23rd, 2024. We believe that \dataset{} is a valuable resource to better understand how people are currently using VLMs and will be the foundation for research in VLM development and evaluation. In the future we plan to continue regular data releases including a large variety of models and multi-image conversations.
\section{Related Works}
\label{sec:related}

\textbf{Crowdsourced Evaluations.} In the past few years, several platforms have emerged that aim to crowdsource evaluation for LLMs and VLMs by allowing users to provide preference votes. These platforms, such as Chatbot Arena~\cite{chiang2024chatbot}, allow anyone to freely engage in open-ended conversations with state-of-the-art commercial and open-source models. Users are able to directly chat with specific models or chat with pairs of anonymous models side-by-side. In the anonymous side-by-side mode, users can provide direct feedback on which responses they preferred, which is used to build a leaderboard. WildVision~\cite{lu2024wildvision} adopts a similar style to Chatbot Arena except that users interact with VLMs instead of LLMs. Our platform builds upon these works by creating a unified interface that allows users to chat with \textit{either} LLMs or VLMs. 

\textbf{Public Chat Datasets:} LMSYS-Chat-1M~\cite{zheng2023lmsyschat1m}, OpenAssistant~\cite{kopf2023openassistant}, and WildChat-1M~\cite{kopf2023openassistant, deng2024wildvisopensourcevisualizer} are all public datasets constructed by capturing users conversations with state-of-the-art LLMs. 
These datasets have been highly influential because they represent more natural human conversations and often contain reward feedback signals. 
However, because these datasets capture text-only conversations,
they do not provide insight into how users incorporate images into conversations and the behavior of VLMs. 
Building on the success of public chat datasets, there is a recent effort to extend the public chat datasets to the visual domain with WildVision, a dataset of 45k chat logs with 9 visual question answering models and 10.4K battle logs across 19 models. In contrast, our \dataset{} data set contains 200K chat logs and 30K battle logs across 40+ models, including all of the strong proprietary models and many open-source models, making it the largest and most complete VLM conversation dataset to date. See Table~\ref{tab:dataset_comparrison} for dataset comparison. 
% It is important to note that while Chatbot Arena does have a VLM leaderboard with over 100,000 votes, they have yet to release any data to the public.

\textbf{VLM Benchmarks:} 
Currently, VLM benchmarks are typically static datasets that have close-ended ground truth answers (either multiple-choice or predefined-string). Some popular examples of these benchmarks include MMMU \cite{yue2024mmmumassivemultidisciplinemultimodal}, DocVQA \cite{mathew2020docvqa}, MME \cite{fu2024mmecomprehensiveevaluationbenchmark}, and VQA 2.0 \cite{goyal2017makingvvqamatter}. To combat against static nature of datasets and minimize test-set contamination, live benchmarks are also available. For example, LiveXiv incorporates updated ArXiv manuscripts for VQA \cite{shabtay2024livexivmultimodallive}. To incorporate benchmarking on open-ended responses, there has been a trend towards using strong models (e.g. GPT-4o) for VLM-as-a-judge to approximate human preference. Some notable examples include WildVision-Bench \cite{lu2024wildvision} and Prometheus-Vision \cite{lee2024prometheusvisionvisionlanguagemodeljudge}. We similarly adopt the VLM-as-a-judge framework to create \dataset{}-Bench, curated from questions from VisionArena, allowing it to be crowdsourced, open, and live. We show that using \dataset{}-Bench, we achieve better correlation and agreement with the VLM leaderboard on Chatbot Arena, which itself has 100x more votes.

\definecolor{gray}{gray}{0.8}
\definecolor{gold}{rgb}{0.85, 0.65, 0.13}
\definecolor{darkbrown}{rgb}{0.6, 0.4, 0.2}
\definecolor{blackish}{rgb}{0.2, 0.2, 0.2}

\section{Dataset and Platform}
\label{sec:dataset}

\subsection{Interface}

VisionArena-Chat and VisionArena-Battle were collected from Chatbot Arena~\cite{chiang2024chatbot}, an open-source platform for evaluating large language models by human preference. On our platform, users are able to directly chat with specific models (direct chat) or chat with pairs of anonymous models side-by-side (battle mode). In battle mode, users can provide direct feedback on which responses they preferred, which is used to build a leaderboard. We refer to these anonymous side-by-side chats as `battles'. An example of our interface can be found in the supplemental. Unlike previous VLM crowdsourcing platforms, we integrate LLMs and VLMs into one unified chat interface with a simple routing mechanism. In side-by-side chat, if a user uploads an image in the first turn of their conversation, we automatically select from two available VLMs; otherwise we sample from the available LLMs. We then collect the votes from the image conversations to compute the VLM leaderboard. We believe this encourages users who may have initially been interested in interacting with LLMs to interact with VLMs as well. In \autoref{sec:analysis}, we show that our conversations do have important distributional differences with WildVision. 

Before using our service, users must accept terms of use, giving us their consent to store and release the conversation data. The platform is free to use, and there is no registration process. We are supported by sponsorships with inference providers. \dataset{} is a subset of conversations collected from February 2024 to September 2024. Given the language and question distribution of our collected conversations (\autoref{fig:lang_battle_counts}, \autoref{fig:category_counts}), the majority of our users are likely located in North America, Europe, and East Asia and work in STEM related fields. 

To encourage user interaction, we provide a `random image' button, which samples from a preset bank of images from 5 datasets: NewYorker~\cite{newyorkernextmldataset}, ChartVQA~\cite{masry-etal-2022-chartqa}, DocVQA~\cite{mathew2020docvqa}, TextVQA~\cite{singh2019towards}, and WikiArt~\cite{wikiart}. We exclude these in VisionArena-Battle as we aim to capture the natural distribution of user inputs when computing leaderboard rankings, but we do include conversations with preset images in VisionArena-Chat, which make up around 15\% of conversations. 

\noindent\textbf{Moderation.} We apply several moderation steps before sending the prompt to the model provider and perform data cleaning procedures before releasing this data to the public.
Before the user receives the response from the model, we perform (not safe for work) NSFW and (child sexual abuse material) CSAM~\cite{PhotoDNA} image detection and then tag and terminate conversations that contain sexual, hateful, or violent content.
For battles, we also perform OpenAI text moderation~\cite{OpenAITextModeration} on user text prompts and discard any responses which contain a violation. For direct chats, we only perform OpenAI text moderation on proprietary models to follow their usage policies. We do not perform text moderation on prompts for open-source models, which opens this data for future analysis. 

Finally, as part of our data release process we use Google’s Vision API~\cite{googlevision} to remove personally identifiable information (PII) from both images and text, removing any content containing human faces or identifiable details.  However, these automated detectors are not infallible, so our dataset may still contain NSFW content or PII. We encourage users who find such instances to notify the authors so the material can be removed.

\subsection{From Preference to Leaderboard Ranking}
\label{sec:leaderboard_calculation}

Using preference votes from pairwise battles in anonymous side-by-side chat, we apply a Bradley-Terry (BT) model~\cite{BradleyTerry1952} to estimate the relative strengths of models through logistic regression. The model’s coefficients serve as \textit{arena scores}, which determine the leaderboard rankings.

Let $n$ denote the number of pairwise comparisons (battles) and $M$ the number of models. 
For each battle $i \in [n]$, we define:
\begin{itemize}
    \item $X_i \in \mathbb{R}^M$: $X_{i,m} = 1$ if model $m$ is presented first to the judge, $X_{i,m} = -1$ if presented last, and 0 otherwise.
    \item  $Y_i \in {0,1}$: The outcome, where 1 indicates the first model won.
\end{itemize}

\begin{figure}[h]
    \centering
    \includegraphics[width=\linewidth, trim=0 30 0 20, clip]{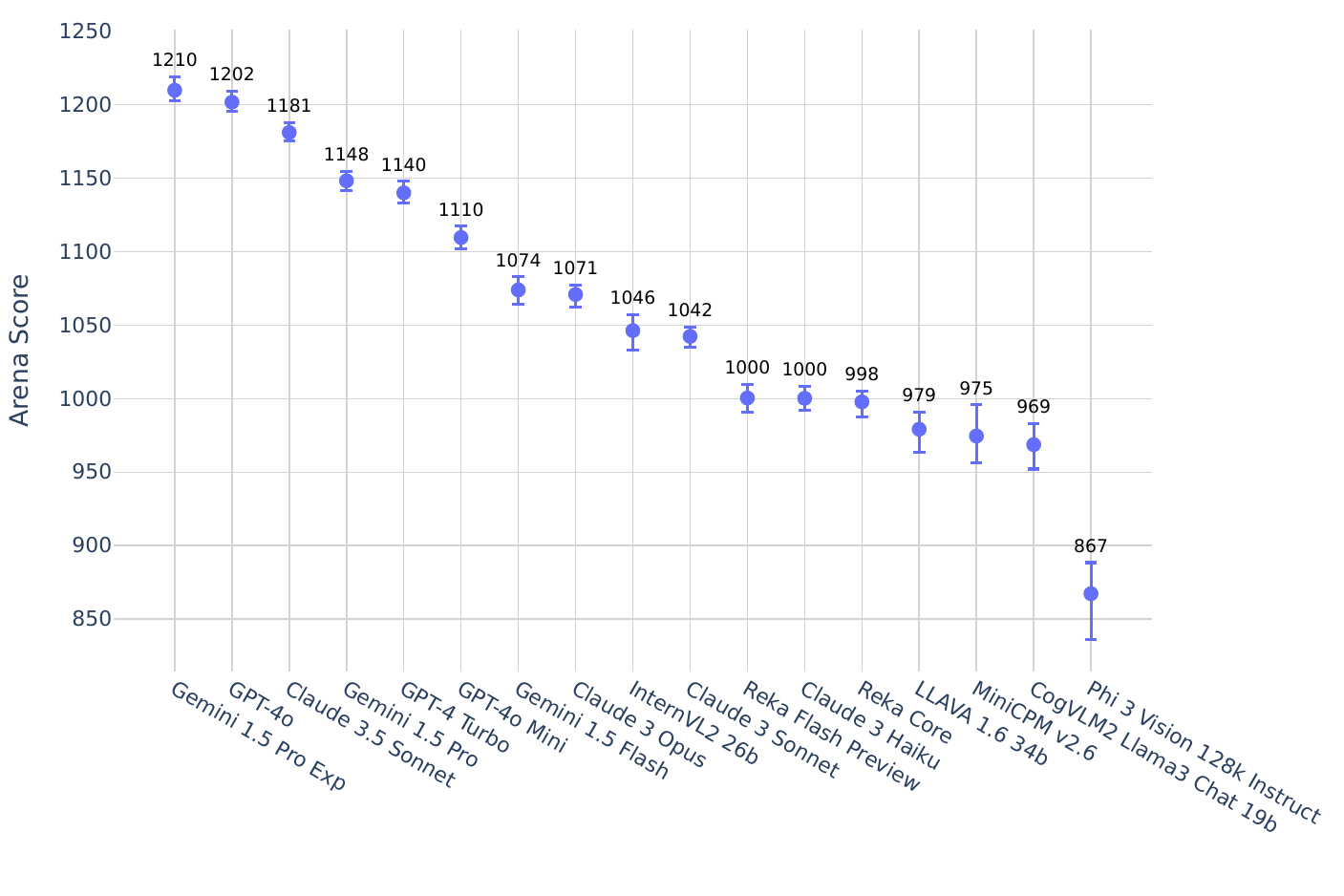}
    \vspace{-1.8em}
    \caption{\textbf{Bootstrap B.T. model scores for VisionArena-Battle.} Proprietary models like Gemini 1.5 Pro and GPT-4o are at the top of the leaderboard, with open models like Llava 1.6, MiniCPM, CogVLMv2, and Phi3 obtaining the lowest ratings. InternVL2 is the highest rated open model, although as shown in Section~\ref{sec:style_analysis}, this is largely due to response style rather than model capability.}
    \label{fig:bootstrap_elo_overall}
\end{figure}

The BT model estimates model strengths $\beta \in \mathbb{R}^M$ through logistic regression:
\begin{equation}
\hat{\beta} = \arg \min_{\beta \in \mathbb{R}^M} \frac{1}{n}\sum\limits_{i=1}^n \text{CE}(\sigma(X_i^\top \beta), Y_i)
\end{equation}
where $\text{CE}$ represents the cross-entropy loss and $\sigma$ is the sigmoid function. 
The BT coefficients $\hat{\beta}$ are the ratings associated with each of the VLMs in the arena.
These BT ratings are used to create the ordered ranking of models on the leaderboard. 
We bootstrap the BT rating estimate 100 times to construct a confidence interval for each rating (\autoref{fig:bootstrap_elo_overall}).

\begin{figure}
    \centering
    \includegraphics[width=\linewidth]{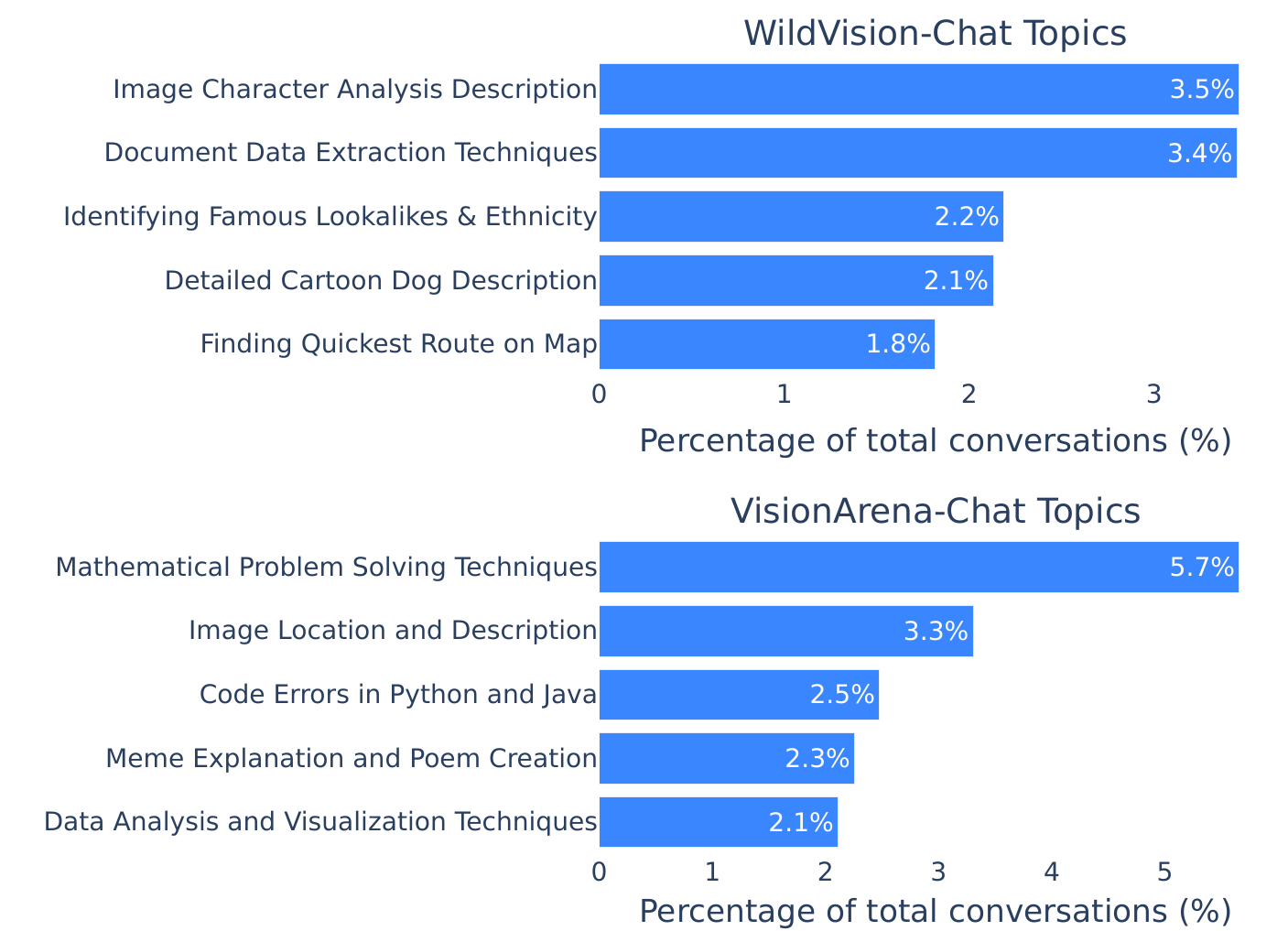}
    \caption{\textbf{Comparison of top 5 topic clusters between WildVision-Chat and VisionArena-Chat.} Compared to WildVision, the most popular topics clusters in \dataset{} capture more real world tasks, specifically in STEM fields.}
    \label{fig:topics_5}
    \vspace{-1em}
\end{figure}

\section{Data Analysis}
\label{sec:analysis}

In the following section, we (1) compare user votes to expert annotators (2), analyze and categorize the distribution of user conversations (3), compute per category leaderboard (4), measure the impact of response style on human preference, and (5) provide examples of difficult questions. 

\subsection{Comparing with Experts and the VisionArena}
To inspect the quality of the human preference votes on our platform, we check their alignment with experts opinions. 
We sample 5 battles in English for each model pair and have 4 experts (PhD Students) label the battles based on their preference, creating an expert-labeled dataset of 516 responses.

We then compute BT scores using the expert labels and compare those scores with the BT scores computed directly on the VisionArena. 
We then compute the Pearson correlation coefficient~\cite{pearson} and the Spearman rank correlation coefficient~\cite{Spearman1904}. 
We obtain a Pearson correlation of 0.88
indicating a strong linear predictive relationship between the BT scores computed by experts and those obtained from the live VisionArena.
We obtain a Spearman rank correlation of 0.87
indicating high agreement in the ordering between the leaderboard rankings and the ranking obtained by our expert labelers on the small subset of data. 

Additionally, for a subset of 100 battles labeled by 3 experts, we observe an agreement of 0.72 (excluding ties) and 0.56 (including ties) between users and expert annotators, compared to 0.77 (excluding ties) and 0.59 (including ties) among expert annotators themselves, further demonstrating the reliability of user votes.

\subsection{What types of questions do people ask?}

We perform topic modeling analysis on the VisionArena-Chat prompts. Following the BERTopic framework, We randomly sample 50K English conversations and embed the documents using CLIP-ViT-B-32, perform DBSCAN clustering, and use GPT-4o to summarize each cluster \cite{grootendorst2022bertopic}. We plot our top 5 clusters in \autoref{fig:topics_5} (the top 20 clusters can be found in the supplement). We find that many people use VLMs to solve math and code problems, identify paintings and geographical locations, perform data analysis on tables and diagrams, explain humorous images, and create stories based on images. Notably, VisionArena-Chat contains important use cases not seen in WildVision-Chat including coding and web UI design problems, handwritten text extraction, and diagram analysis. Manually inspecting the clusters, we also see that WildVision-Chat's clusters are often very specific to a certain task (e.g. "Detailed Cartoon Dog Description", "Rice Leaf Disease Identification"), while VisionArena-Chat's cluster descriptions are broader which indicates the diversity of our data. Surprisingly, the majority of our questions require OCR, and we receive a large number of homework problems and diagram understanding questions. In the following section, we construct categories for each of these major use cases. 

\subsection{Prompt Categories}

\begin{figure}[h]
    \centering
    \begin{minipage}[t]{0.48\textwidth}
        \centering
        \small
        \begin{tabularx}{\textwidth}{lX}
            \toprule
            \textbf{Category} & \textbf{Description} \\
            \midrule
            \textit{Multi-Turn} & Conversations with multiple turns. \\
            \textit{Exclude Ties} & Battles which do not end in a tie. \\
            \textit{Exclude Refusal} & Neither model refuses to answer. \\
            \midrule
            \textit{Captioning} & \textit{Only} asks for a description of the image. \\
            \textit{OCR} & Requires reading text within the image. \\
            \textit{Coding} & Contains a code block in either the user inputs or model outputs. \\
            \textit{Entity Recognition} & Asks to identify objects, places, or people in the image. \\
            \textit{Homework} & Requires answering a problem which likely comes from a homework or exam. \\
            \textit{Humor} & Asks to explain the humor within the image or ask for a humorous composition. \\
            \textit{Diagram} & Contains images with a diagram (e.g., flowchart, circuit, graph). \\
            \textit{Creative Writing} & Asks for a creative composition such as a story or a script. \\
            \bottomrule
        \end{tabularx}
        \vspace{-1em}
        \caption{Descriptions of \dataset{} categories.}
        \vspace{3em}
        \label{tab:dataset_categories}
    \end{minipage}
    \hfill
    \begin{minipage}[t]{0.48\textwidth}
        \centering
        \includegraphics[width=\linewidth]{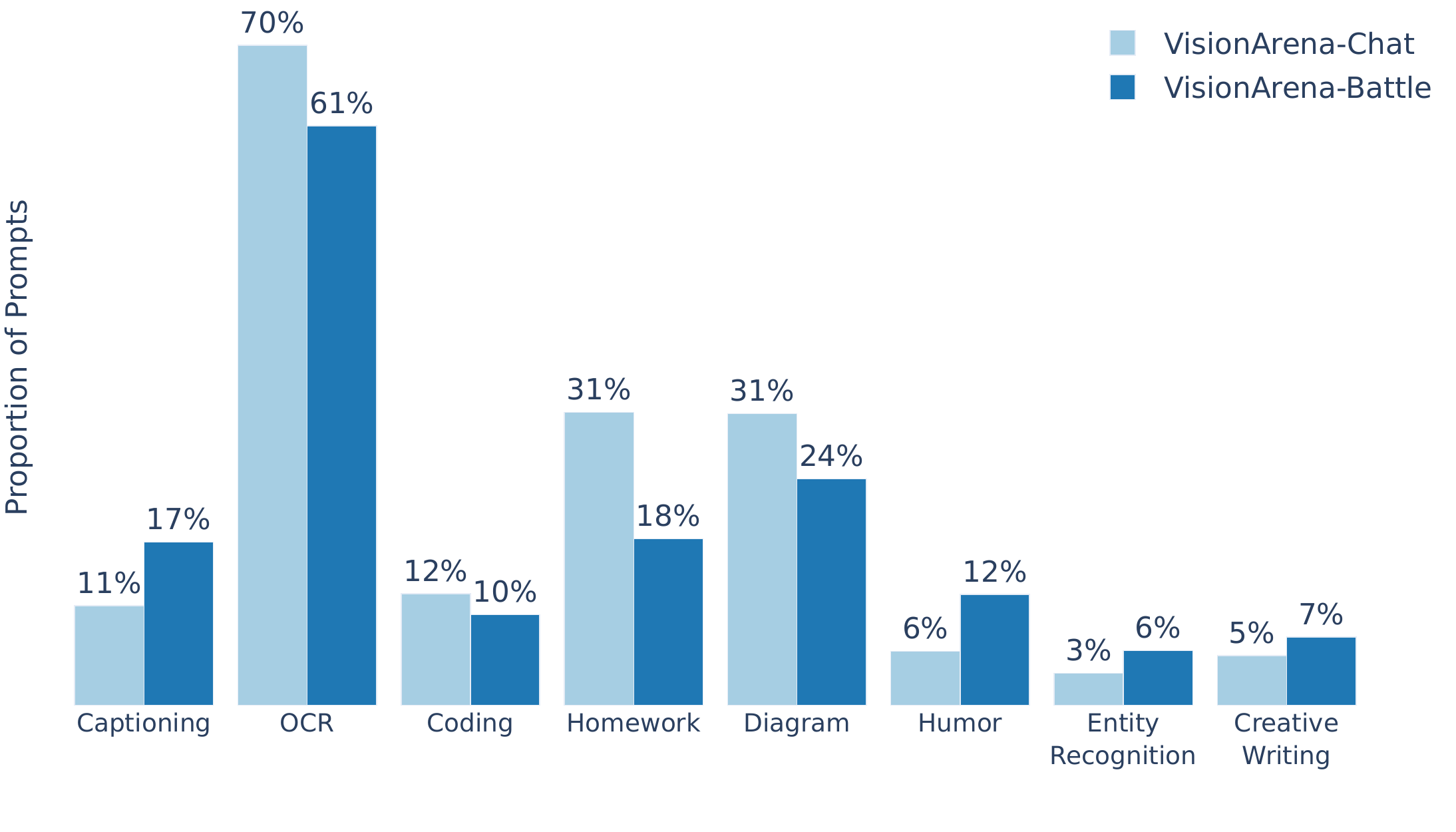}
        \vspace{-3em}
        \caption{\textbf{Category Distribution.} Excluding preset examples. We see that direct chat data contains a higher proportion of coding, homework, and diagram questions while battle data contains more captioning, humor, and creative writing questions.}
        \label{fig:category_counts}
        % \vspace{-1em}
    \end{minipage}
    \hfill
    \begin{minipage}[t]{0.48\textwidth}
        \centering
        \vspace{2em}
        \includegraphics[width=\linewidth, trim=0 0 0 0, clip]{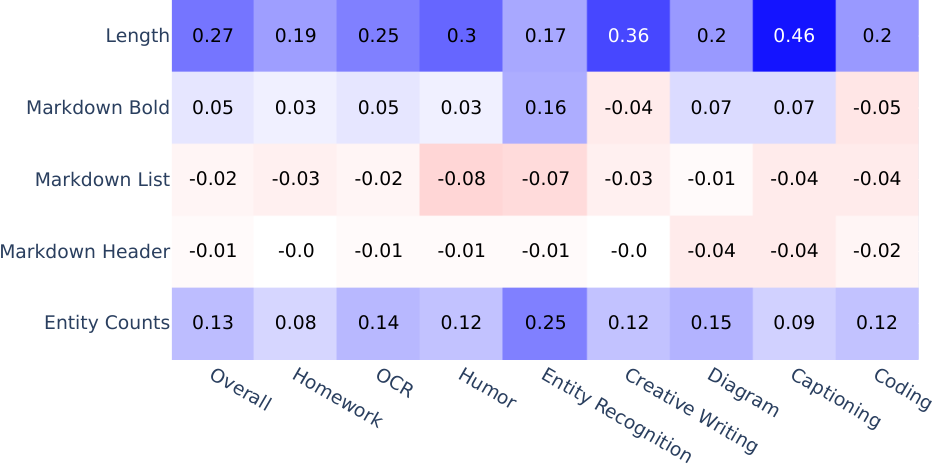}
        \caption{\textbf{Impact of confounding variables on user preferences, measured by $\hat{\gamma}$ in the enhanced Bradley-Terry Model.} Length is by far the most influential stylistic factor, with higher influence on preference for more open ended questions like humor, creative writing, and captioning.}
        \label{fig:control_coefs}
        \vspace{-2em}
    \end{minipage}
\end{figure}

\begin{figure*}[h]
    \includegraphics[width=\linewidth]{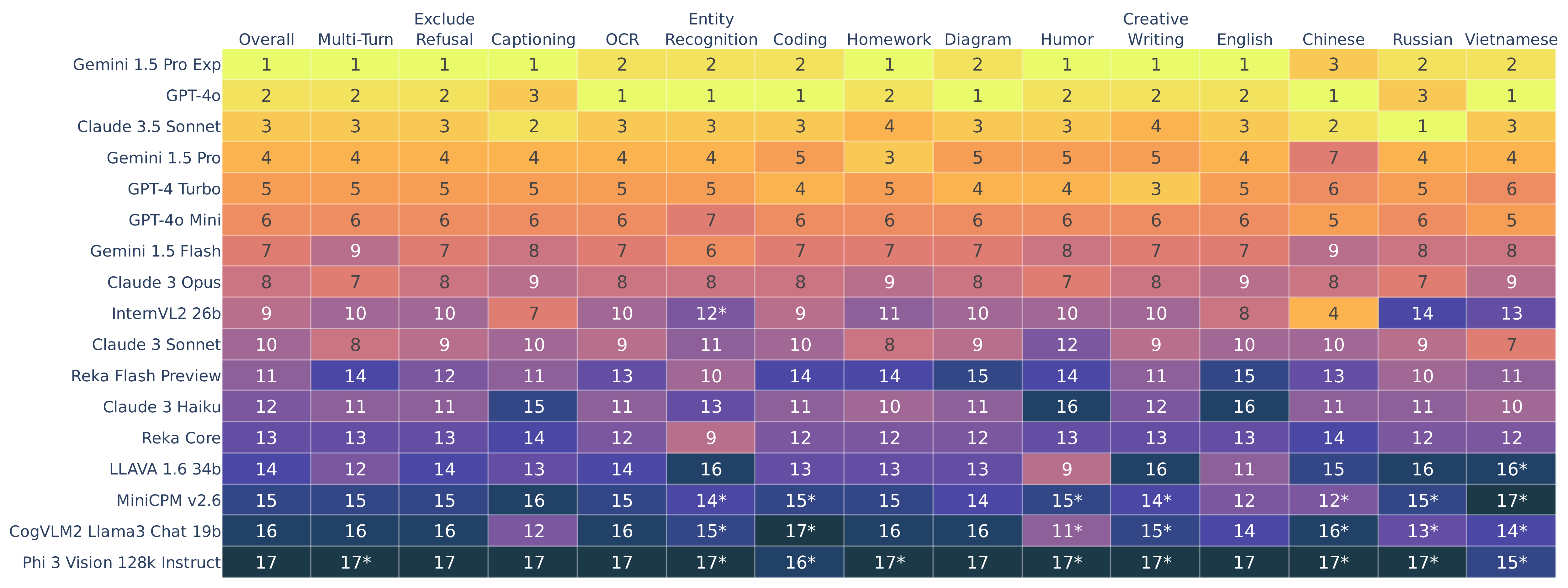}
    \label{fig:category_leaderboard}
\vspace{-2em}
\caption{\textbf{Model rankings across question categories and languages.} Cells with * have fewer than 100 votes. Certain models achieve a much higher ranking for a particular category, such as InternVL2 on captioning and Chinese, Reka Core on entity recognition, and Llava 1.6 on Humor. Conversely, we see certain model rankings drop, such as Reka Flash on multi-turn and diagrams, and Gemini on Chinese.}
\vspace{-0.3em}
\label{fig:category_battle_results}
\end{figure*}

Based on the clustering analysis and manual inspection, we manually define 8 non-disjoint categories that reflect the vast majority of prompts and test different capabilities of the VLM.
We use Gemini 1.5 Flash to classify each prompt, using both the image and text, into a set of predefined categories listed in \autoref{tab:dataset_categories}.
Section~\ref{sec:category_supp} contains detailed descriptions of each category, the prompts used to implement the categorization with 1.5 Flash, and the correlation between 1.5 Flash category labels and those of SOTA models.

\begin{figure*}
\includegraphics[width=\linewidth, trim=0 70 0 80, clip]{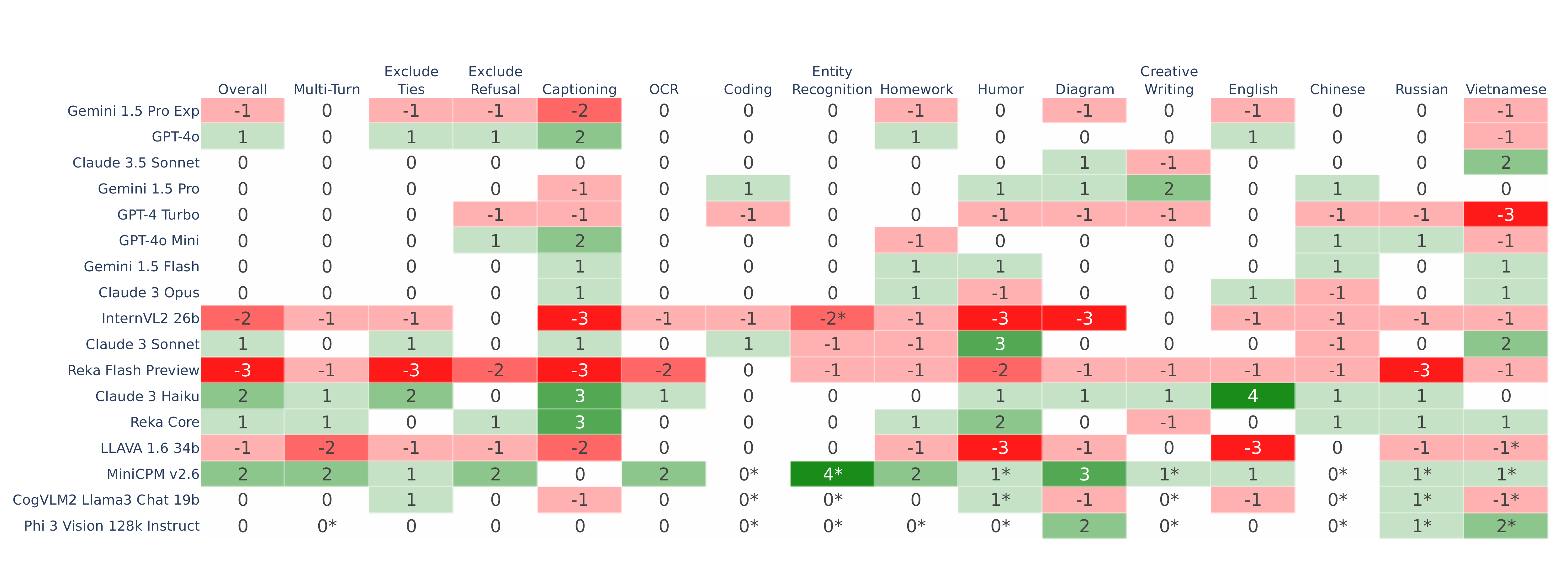}
\vspace{-1.8em}
\caption{\textbf{Change in model rankings when style control is applied.} Cells with * contain battles with fewer than 100 votes. InternVL, Reka Flash, and Llava 1.6 see a ranking drop across the majority of categories, while Claude 3 Haiku and MiniCPM see an increase across most categories. The model lineup changes the most for the captioning, humor, and Vietnamese categories.}
\vspace{-1em}
\label{fig:style_control}
\end{figure*}

\autoref{fig:category_counts} shows the distribution of category counts for both battles and direct chat conversations. We observe that direct chat conversations contain more homework and diagram understanding problems while battles contain more humor, captioning, and creative writing problems. We speculate this is because users in direct chat mode are more interested in using proprietary VLMs at no cost to assist them with their daily tasks. 

Model rankings and Arena Scores for select categories and languages are shown in \autoref{fig:category_battle_results} (full table and arena scores can be found in the supplemental). We find several interesting insights such as: 

\begin{itemize}
    \item Gemini 1.5 Pro Exp~\cite{reid2024gemini}, InternVL~\cite{chen2024far,chen2023internvl} and Reka Flash~\cite{rekateam2024rekacoreflashedge} achieve a worse ranking for categories with require OCR like coding, homework, and diagrams. 
    \item InternVL has a large improvement in ranking for the captioning category. In the following section, we show that this is largely due to stylistic choices such as formatting. 
    \item Claude Opus, Sonnet, and Haiku~\cite{claude32024family} see an increase in ratings for multi-turn conversations.
    \item The Gemini class of models drops in performance on non-english conversations.
\end{itemize}

\subsection{Controlling stylistic biases in evaluations}
\label{sec:style_analysis}

The Vision Arena captures signals from users of various backgrounds and preferences to construct its leaderboard. However, recent literature has pointed out potential confounding variables in model evaluation such as the length of the response or stylistic formatting \cite{chiang2024chatbot, dunlap_vibecheck}. Others have also mentioned various axes in which annotators may disagree including task underspecification, response style, refusals, and annotation errors \cite{zhang2024divergingpreferencesannotatorsdisagree}. Thus, we explore the effect of these stylistic features on the VisionArena user preference. 

We follow recent work that extends the BT model to include style features \cite{li2024crowdsourceddatahighqualitybenchmarks}. Given a set of style features (e.g. response length, number of markdown headers), we add a style vector to the BT model $\vec{Z}$ where=
    $Z_i \in \mathbb{R}^S$ is a vector of $S$ style features.
the enhanced BT model has the style coefficients $\gamma \in \mathbb{R}^{S}$:
\begin{align*}
    \hat{\beta}, \hat{\gamma} = \arg \min_{\beta \in \mathbb{R}^M, \gamma \in \mathbb{R}^S} \frac{1}{n}\sum\limits_{i=1}^n \text{CE}(\sigma(X_i^\top \beta + Z_i^\top \gamma), Y_i)
\end{align*}
 For each style feature $Z_i$, we compute the normalized difference between the feature values of both model responses. 
 The resulting $\hat{\beta}$ represents model strengths adjusted for style effects, while $\hat{\gamma}$ quantifies the influence of style on user preferences.
 
To control for these stylistic factors, we modify how \dataset{} computes the model scores by accounting for the stylistic differences between two answers (response length, number of markdown headers, etc) as additional features to the existing BT model.

\noindent\textbf{Controlling for length and markdown.} Applying style control, we see the captioning category is heavily affected by style, with \autoref{fig:style_control} showing a large difference in model rankings. We suspect this is because VLMs are heavily optimized for captioning and can usually correctly identify the main subjects and context of the image. This is supported in \autoref{tab:bothbad_counts}, which shows captioning questions have the smallest proportion of `both bad' votes. In cases where both models provide a reasonable description of the image, the user may rely on stylistic features to determine preference.

Furthermore, models like InternVL and Reka Flash Preview see a large decrease in rankings when style control is applied. \autoref{fig:token_count} shows these models have an unusually high output token count compared to models of a similar class, indicating that some models may be `preference hacking' by training their models to produce long or nicely-formatted outputs. While this is not necessarily bad, it is important to consider when decoupling preference from capability. 

\begin{table}[h]
    \centering
    \resizebox{0.48\textwidth}{!}{%
    \begin{tabular}{cccccccc}
    \toprule
     Caption & Homework & OCR & Coding & Humor & \shortstack{Entity \\ Recog.} & \shortstack{Creative \\ Writing} & Diagram \\
    \midrule
    11.33 & 31.12 & 20.62 & 29.4 & 18.47 & 22.25 & 14.21 & 26.06 \\
    \bottomrule
    \end{tabular}%
    }
    \vspace{-0.5em}
    \caption{\textbf{Percentage of `tie (bothbad)' per category.} Captioning, creative writing, and humor categories have low percentages of bothbad responses compared to coding and homework problems.}
    \label{tab:bothbad_counts}
    \vspace{-0.6em}
\end{table}

% \begin{figure}[h]
%     \centering
%     \includegraphics[width=\linewidth, trim=0 0 0 0, clip]{figs/style_coeffs.pdf}
%     \caption{\textbf{Impact of confounding variables on user preferences, measured by $\hat{\gamma}$ in the enhanced Bradley-Terry Model.} Length is by far the most influential stylistic factor, with higher influence on preference for more open ended questions like humor, creative writing, and captioning.}
%     \label{fig:control_coefs}
% \end{figure}

\begin{figure*}[!htb]
    \centering
    \begin{subfigure}{0.49\linewidth}
        \centering
        \includegraphics[width=\linewidth, trim=0 0 0 0, clip]{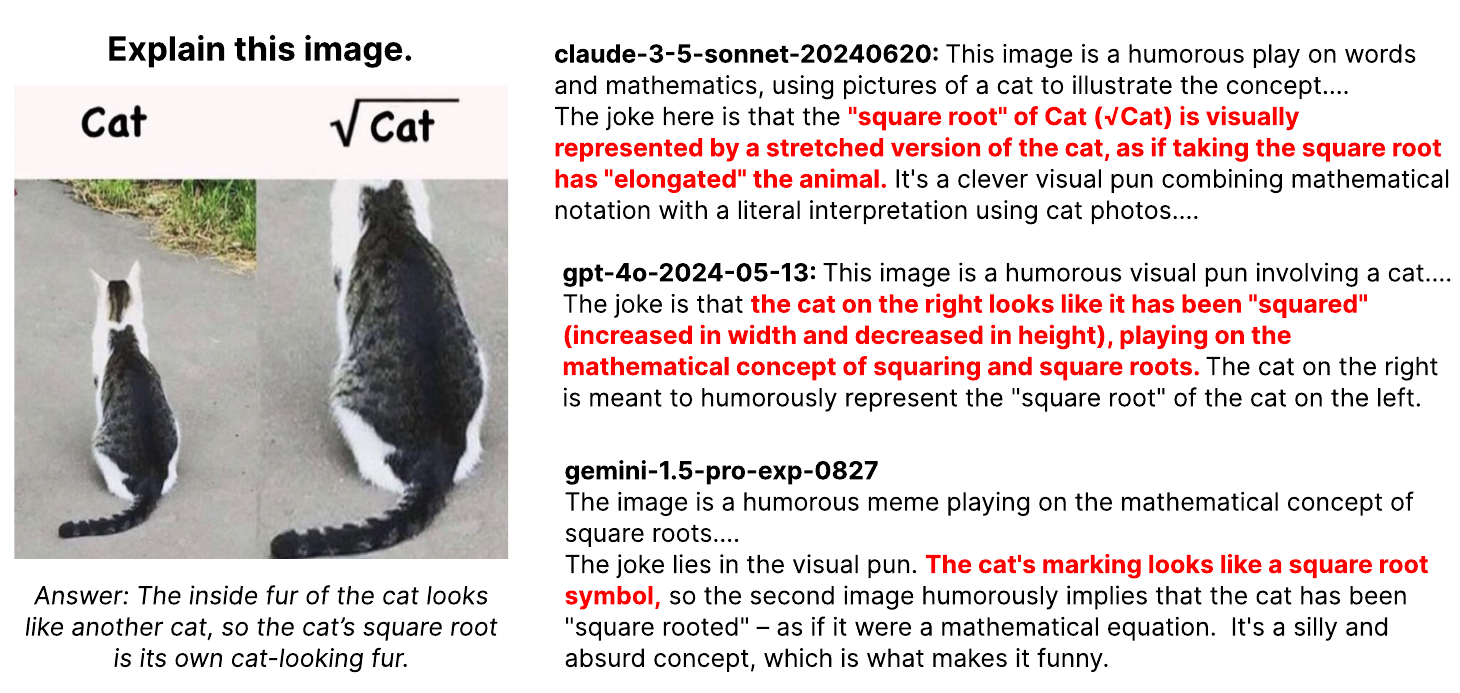}
        \caption{\textbf{Failure case 1.}}
        \label{fig:cat_example}
    \end{subfigure}
    \hfill
    \begin{subfigure}{0.49\linewidth}
        \centering
        \includegraphics[width=\linewidth, trim=0 0 0 0, clip]{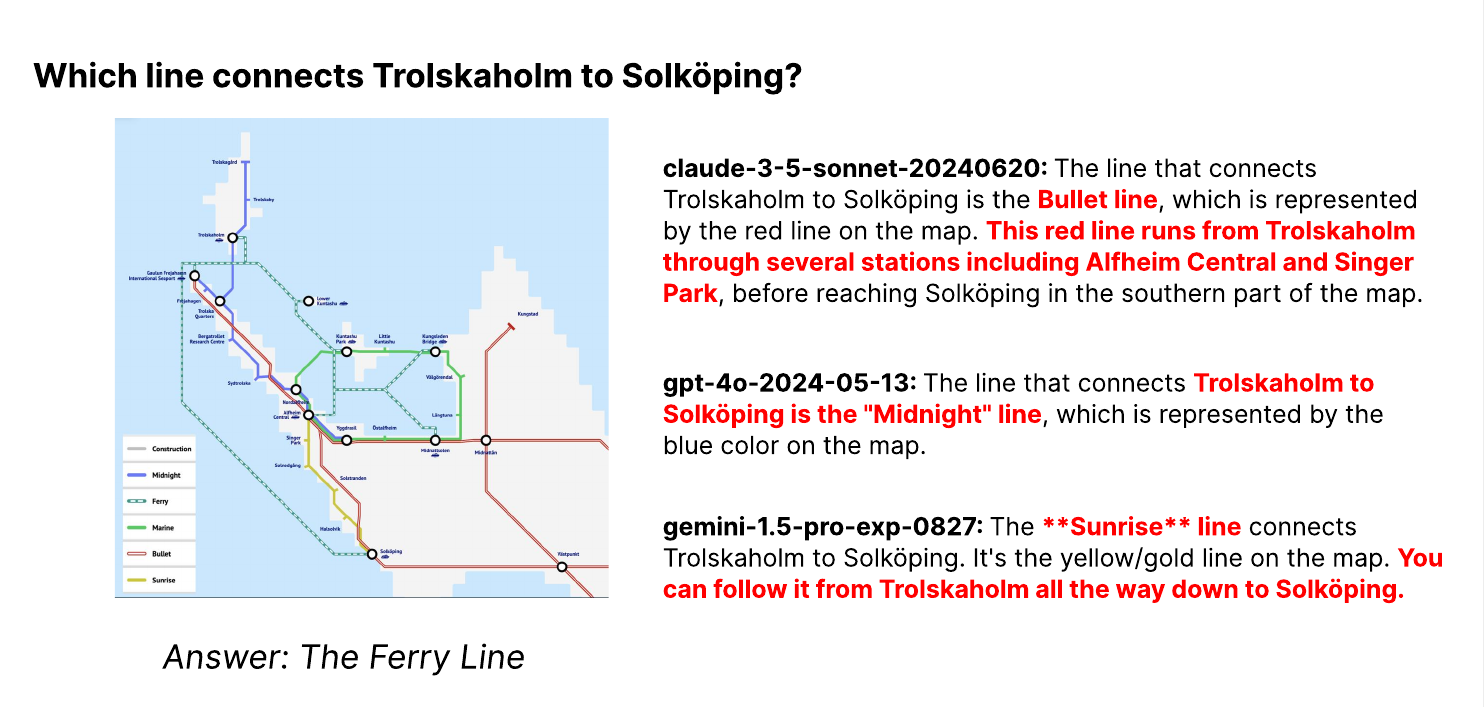}
        \caption{\textbf{Failure case 2.}}
        \label{fig:train_line_example}
    \end{subfigure}
    \vspace{-0.5em}
    \caption{\textbf{VLM failure modes.}  The top proprietary models fail on questions which require advanced visual reasoning. For example, failure case 1 requires the visual understanding that the cat's fur patterns looks like another smaller black cat, and the linguistic connection between this and a square root.  
    Non-truncated outputs in Section~\ref{sec:failures_supp}.} 
    \vspace{-1.8em}
    \label{fig:failures}
\end{figure*}

\noindent\textbf{Controlling for specificity.} We further extend the Bradley-Terry model to include the effect of response specificity.
We define the complexity of a response as the number of named entities in the response. 
We use a NER model \cite{NOTHMAN2013151} to tag each response and use the number of named entities as our specificity score. In \autoref{fig:control_coefs}, we see that users prefer high specificity for tasks like entity recognition and diagram understanding while placing less emphasis on specificity for tasks like captioning and homework.

\subsection{Failure Cases}

We use \dataset{} to analyze examples which are particularly challenging for current VLMs.
% While VLMs have served as an incredible resource for many users on our platform, there are many simple cases where they fail. We demonstrate how the dataset can be used to find cases where models fail and where humans would succeed.
We first filter VisionArena-Chat for prompts where the user voted that both models are bad, and collect a pool of 10 images which most or all of the current VLMs fail. The full set can be found in the supplemental section.
% We divide failure cases into three categories: hard OCR, counting, and reasoning. Lastly, we provide examples of hard examples on which most or all of the current VLM's fail, with additional examples in Section~\ref{sec:failures_supp}.
\autoref{fig:failures} shows two examples of user questions that require advanced visual reasoning. This example requires the model to (1) understand that the two cats in the image are the same (2) the pattern on the cats back gives the illusion of another cat (3) this illusion of a smaller cat in a cat can be related to the fact that the square root of a number is a smaller demonimation of that number. \autoref{fig:train_line_example} is an example of fine-grained spatial understanding, as the model must locate both locations and reason over the many intersecting lines in the image. Analysis of other failure cases in the supplemental section indicates that current VLMs still struggle on visual grounding tasks like reading distorted images, spatial understand and counting, as well as more complex reasoning tasks. 
% Additionally, when looking at the cross section of categories with the highest rates of 'both bad' votes, we see that homework problems 

% \section{Applications} 
% \label{sec:applications}

% \begin{table}[ht]
% \centering
% \begin{tabular}{lccc}
% \tiny
% \toprule
% \textbf{Benchmark} & \textbf{WV-Bench} &  \textbf{ArenaVQA} & \textbf{Arena-Hard-VQA} \\
% \midrule
% Confidence Agreement & 89.1\% & 66.7\% & 84.8\%  \\
% Separability         & 87.4\% & 83.68\% & 82.11\%  \\
% Spearman Correlation & 94.1\% & 77.0\% & 95.2\%  \\
% \bottomrule
% \end{tabular}
% \caption{Table comparing various benchmarks on different evaluation metrics. \textbf{\textcolor{red}{TODO: FILL THIS OUT}}}
% \end{table}

\section{Instruction tuning vision-language models}
\label{sec:finetuning}

Effective instruction finetuning for vision-language models depends on the diversity of instructions, the difficulty of prompts, and the quality of responses. This section demonstrates the potential of \dataset{} for training high-performance instruction-following models.

\footnotetext[1]{gpt-4o-mini-2024-07-18, gpt-4-turbo-2024-04-09, gemini-1.5-pro-api-0514, claude-3-5-sonnet-20240620, gemini-1.5-pro-exp-0827, gpt-4o-2024-05-13, gemini-1.5-pro-exp-0801, gemini-1.5-flash-api-0514, claude-3-opus-20240229, gemini-1.5-flash-exp-0827, gemini-1.5-flash-8b-exp-0827, llama-3.2-vision-90b-instruct, qwen2-vl-72b \cite{wang2024qwen2vlenhancingvisionlanguagemodels}, gpt-4o-2024-08-06, chatgpt-4o-latest-20240903, chatgpt-4o-latest-20240808}

We curate a high quality instruction-tuning dataset by sampling from the conversations with the highest-performing VLMs. We choose 100,000 conversations from VisionArena-Chat from the top models. This led to a dataset with conversations from 16 different models\footnotemark[1] including proprietary models such as GPT-4o~\cite{openai2023gpt4}, Gemini-1.5-Pro~\cite{reid2024gemini},  and Claude-3.5-Sonnet~\cite{claude32024family} as well as open-source models such as Qwen2-VL-72B~\cite{Qwen-VL, wang2024qwen2vlenhancingvisionlanguagemodels} and Llama-3.2-90B-Vision-Instruct~\cite{llama3modelcard}. We compare the effectiveness of this dataset for finetuning to a 100K subset of the Llava-Instruct-158K~\cite{liu2023visualinstructiontuning}.

We use Llama-3.2-11B-Vision and freeze the vision encoder while finetuning the multimodal projector and language model. We finetune for 3 epochs on the data for both our 100k dataset and the 100k Llava-Instruct dataset. In \autoref{table:finetune}, we label the model finetuned on our \dataset{} data Llama-3.2-VisionArena and the model trained using Llava-Instruct-158K Llama-3.2-Llava-Instruct.

The evaluation results are shown in \autoref{table:finetune}. Llama-3.2-VisionArena significantly outperforms Llama-3.2-Llava-Instruct on MME, HallusionBench, MMMU, MMMU-Pro, and WildVision-Bench. Additionally, Llama-3.2-VisionArena outperforms Llama-3.2-11B-Vision-Instruct on both MME and WV-Bench, despite being fine-tuned on 30X less data. In \autoref{sec:contamination_supp}, we show that these improvements are not due to contamination of these benchmarks 
% Additionally, increases the MMMU baseline by 1.3 points compared to the base model. 
% This demonstrates the effectiveness of VisionArena to improve the capabilities of VLM's.

% \begin{table}[h]
% \small
% \centering
% \resizebox{0.47\textwidth}{!}{%
% \begin{tabular}{lccc}
% \toprule
% \textbf{Model} & \textbf{Samples} & \textbf{MMMU}\cite{yue2024mmmumassivemultidisciplinemultimodal} & \textbf{WV-Bench} \\
% \midrule
% Llama3.2-11B-V & - & 41.7 & - \\
% \textcolor{darkgray}{Llama3.2-11B-V-Instruct} & \textcolor{darkgray}{3M+} & \textcolor{darkgray}{\textbf{50.7}} & \textcolor{darkgray}{47.2} \\
% Llama-3.2-Llava-Instruct & 100K & 27.9 & 10.4 \\
% Llama-3.2-VisionArena & 100K & 43.0 & \textbf{56.9} \\
% \bottomrule
% \end{tabular}%
% }
% \caption{\textbf{Performance across models trained with different instruction tuning datasets.}  Scores for Llama-3.2-11B-Vision and Llama-3.2-11B-Vision-Instruct are author-reported. Finetuning with 100k samples from \dataset{} achieves an MMMU score 15 points higher than finetuning on Llava-Instruct data and a WV-Bench score 9 points higher than Llama-3.2-11B-Vision-Instruct. Despite being fine-tuned on 30x less data, Arena-11B performs comparably to Llama-3.2-11B-Vision-Instruct in chat tasks.}
% \label{table:finetune}
% \vspace{-1em}
% \end{table}

\begin{table*}[h]
\vspace{-1em}  % adjust vertical space if needed
\centering
\vspace{-1em}
\resizebox{\textwidth}{!}{%
\begin{tabular}{lccccccccc}
\toprule
 & & \multicolumn{2}{c}{\textbf{MME}\cite{fu2024mmecomprehensiveevaluationbenchmark}} & \multicolumn{3}{c}{\textbf{HallusionBench}\cite{Guan_2024_CVPR}} & \textbf{MMMU}\cite{yue2024mmmumassivemultidisciplinemultimodal} & \textbf{MMMU-Pro}\cite{yue2024mmmupro} & \textbf{WV-Bench}\cite{lu2024wildvision}\\
\cmidrule(r){3-4} \cmidrule(r){5-7}
\textbf{Model} & \# Samples & Cog. & Perc. & Acc.(all) & Fig. & Q. & Acc. & Acc. & Acc. \\
\midrule
Llama3.2-11B-V-Instruct & 3M+ & \underline{327.5} & \underline{1421.7} & \textbf{48.6} & \textbf{26.0} & \textbf{23.1} & \textbf{50.7} & \textbf{0.28} & \underline{47.2}\\
Llama-3.2-Llava-Instruct & 100K & 262.1 & 1067.6 & 38.7 & 13.9 & 8.6  & 27.9& 0.12 & 10.4 \\
Llama-3.2-VisionArena & 100K & \textbf{345.4} & \textbf{1437.0} & \underline{45.2} & \underline{19.9} & \underline{16.3} & \underline{43.0} & \underline{0.27} & \textbf{56.9}\\
\bottomrule
\end{tabular}%
}
\vspace{-0.8em}
\caption{\textbf{Performance across models trained with different instruction tuning datasets.}  Llama-3.2-11B-Vision-Instruct scores are author-reported. Fine-tuning on samples from \dataset{}-Chat outperforms fine-tuning on Llava-Instruct across all benchmarks. Llama-3.2-\dataset{} also outperforms Llama-3.2-11B-Vision-Instruct on both MME and WV-Bench, despite being fine-tuned on 30x less data.}
\label{table:finetune}
\vspace{-1.2em}
\end{table*}

\section{VisionArena-Bench: An Automatic Offline Human-Preference Benchmark for VLMs}

Lastly, we demonstrate \dataset{}'s ability to cheaply approximate model preference rankings with \dataset{}-Bench. Currently, online preference benchmarks like Chatbot Arena obtain a ranking for a new model by adding it to their platform and waiting days or weeks to collect enough votes for a stable ranking. For a single developer hoping to test a particular version of their model, obtaining these online model rankings is infeasible.

\footnotetext[2]{gemini-1.5-pro-exp-0827, gemini-1.5-flash-exp-0827\cite{reid2024gemini}, gpt-4o-2024-05-13, gemini-1.5-flash-8b-exp-0827, internvl2-26b \cite{chen2024far}, claude-3.5-sonnet-20240620 \cite{claude32024family}, gpt-4-turbo-2024-04-09 \cite{openai2023gpt4}, claude-3-sonnet-20240229, llama-3.2-11b-vision-instruct \cite{llama3modelcard}, gemini-1.5-pro-001, internvl2-4b, gpt-4o-mini-2024-07-18, claude-3-opus-20240229, gemini-1.5-flash-001, reka-core-20240501 \cite{rekateam2024rekacoreflashedge}, claude-3-haiku-20240307}
% Current benchmarks are static and based on ground-truth answers such as multiple choice, which does not reflect how users use VLMs on real-world tasks.While the Chatbot Arena VLM Leaderboard addresses these issues by recording human preference on battles, collecting sufficient votes may take days or weeks to estimate the model ranking properly. 
We develop a solution for those who need a quick and cheap evaluation of their models: VisionArena-Bench. VisionArena-Bench is a set of 500 diverse image and text prompts that accurately approximates the model ranking from the Chatbot Arena VLM Leaderboard. 

\noindent\textit{Offline benchmark curation.} To gather questions for this offline benchmark, we build on recent work in building benchmarks from crowd-sourced evaluations in LLM and adapt them to the context of VLM ~\cite{li2024crowdsourceddatahighqualitybenchmarks}. We first filter the data for single turn to prevent the user from correcting the model in its response on its second turn. We then filter out non-english conversations as we are personally unable to verify the quality of non-english prompts.

To sample diverse questions, we perform topic modeling using the library BERTopic with multimodal embeddings \cite{grootendorst2022bertopic}. We extract image embeddings and text embeddings using a CLIP model (e.g. CLIP-ViT-B-32) \cite{radford2021learningtransferablevisualmodels}. We then average the image and text embeddings so that each document corresponds to a single embedding. Then, we use UMAP to reduce the dimensions of the embedding and use hierarchical-based clustering (HDBSCAN) to generate topic clusters \cite{mcinnes2020umapuniformmanifoldapproximation, Malzer_2020}. We then uniformly sample from each topic cluster to generate the 500 prompts.

\noindent\textit{Automatic evaluation with VLM-as-a-judge.} To evaluate a model, we use the LLM-as-a-judge framework mentioned in \cite{li2024crowdsourceddatahighqualitybenchmarks} applied to VLMs. We first select a fixed anchor model (GPT-4-Turbo\cite{openai2023gpt4}) that will be used in the pairwise comparisons. To evaluate a given model $M$ on a user prompt $p$, we generate responses for both $M$ and the anchor model on $p$ and then utilize GPT-4o as a judge to provide a preference score between the ($M$, anchor) pair on a 5-point Likert scale. 1 indicates a strong preference for model A and 5 indicates a strong preference for model B. We then obtain this score for all models across all prompts in VisionArena-Bench to obtain VLM-generated pairwise preference votes and use the same procedure as described in \autoref{sec:dataset} to produce final model scores.
% We evaluate a model on a given image and prompt using a pairwise comparison against a strong baseline model GPT-4-Turbo \cite{openai2023gpt4}. We then utilize GPT-4o to score each output by rating its preference between the pair on a 5-point Likert scale. 1 indicates a strong preference for model A and 5 indicates a strong preference for model B, which effectively penalizes the model for significant losses. 
In \autoref{sec:vision_arena_bench_supp}, we provide the detailed judge prompt template. To avoid potential bias, we prompt the judge model to judge twice, swapping the response position between the two rounds. 

To evaluate the effectiveness of our benchmark to existing work~\cite{lu2024wildvision}, we leverage standard metrics such as Spearman correlation and Kendall Tau correlation which measure the agreement between two benchmarks' model rankings. We choose a shared set of 16 models and compare the offline benchmark rankings to the online Chatbot Arena. \autoref{table:bench_results} shows
VisionArena-Bench achieves a higher Spearman and Kendall Tau correlation than WildVision-Bench, with a 17.1\% and 20.5\% gain respectively. 
% that our simple benchmark creation strategy achieves a Spearman correlation of 97.3\% and Kendall Tau correlation of 89.7\%, which is higher than other VLM benchmarks like WildVision-Bench by 17.1\% and 20.5\%, respectively. 
In Section~\ref{sec:vision_arena_bench_supp}, we also compare the results using the same baseline model as WildVision (e.g. Claude-3-Sonnet-20240229), showing that the spearman correlation to Chatbot Arena's VLM leaderboard (10/23/2024) remains the same. This demonstrates the potential of VisionArena-Bench as a cost-effective and scalable offline benchmark that closely mirrors human preferences captured in online evaluations, enabling researchers to efficiently assess and compare VLMs without the need for extensive user studies.

\begin{table}[h]
\small
\centering
\begin{tabular}{lcc}
\toprule
\textbf{} & \textbf{VisionArena-Bench} & \textbf{WV-Bench}  \\
\midrule
Confidence Agreement & \textbf{98.6\%} & 87.6\% \\
Spearman Correlation & \textbf{97.3\%} & 80.2\% \\
Kendall Tau Correlation & \textbf{89.7\%} & 69.2\% \\
\bottomrule
\end{tabular}
\vspace{-1em}
\caption{\textbf{Correlation of rankings with ChatbotArena's VLM leaderboard.} Performance comparison on 16 models\protect\footnotemark[2] between VisionArena-Bench to Chatbot Arena's VLM leaderboard based on confidence agreement, spearman correlation, and Kendall tau correlation. Result as of leaderboard on October 23, 2024.}
\vspace{-2em}
\label{table:bench_results}
\end{table}
    \section{Discussion, Limitations, and Future Work}
\label{sec:conclusion}

Human preference benchmarks provide a critical lens for assessing performance on open-ended tasks where an explicit notion of correctness is either unavailable or subjective. Users may implicitly consider factual accuracy when making their preferences, but we would like to emphasize that \textbf{this benchmark is designed to measure human preferences rather than explicitly evaluate factual accuracy.} We see \dataset{} as complementary to existing datasets and benchmarks that measure objective correctness. 

Despite the breadth of coverage offered by \dataset{}, significant gaps remain in representing the full distribution of real-world use cases for vision-language models. As highlighted in \autoref{sec:analysis}, our dataset contains many sample from domains such as STEM problems, OCR tasks, and toy problems (e.g., humor and riddles). These areas, while valuable, leave critical application domains underrepresented, including geospatial applications, medical domains, and visual assistance. Furthermore, while \dataset{} contains over 100 languages, many of these languages do not contain enough examples to produce a stable leaderboard. Looking forward, we hope to enourage a more diverse user base by changing our UI to be multi-lingual and improving general user experience. Lastly, we have made it easy for the community to contribute new question categories and models at \url{https://github.com/lm-sys/FastChat}.

\newpage

{
    \small
    \bibliographystyle{plainnat}
    \bibliography{main}
}

\clearpage
\setcounter{page}{1}
\maketitlesupplementary

\section{Acknowledgments}

We extend our gratitude to Lianmin Zheng and Ying Sheng for their early discussions and implementation of the vision arena in the LMArena platform. We also thank Tianle Li and Anastasios Angelopoulos for their valuable insights and feedback on both the platform and the manuscript. Most of all, we appreciate the community for their contributions through questions and votes, as this platform and dataset would not be possible without their participation.

\begin{figure*}[h]
    \centering
    \vspace{2em}
    \includegraphics[width=\linewidth, trim=0 0 0 0, clip]{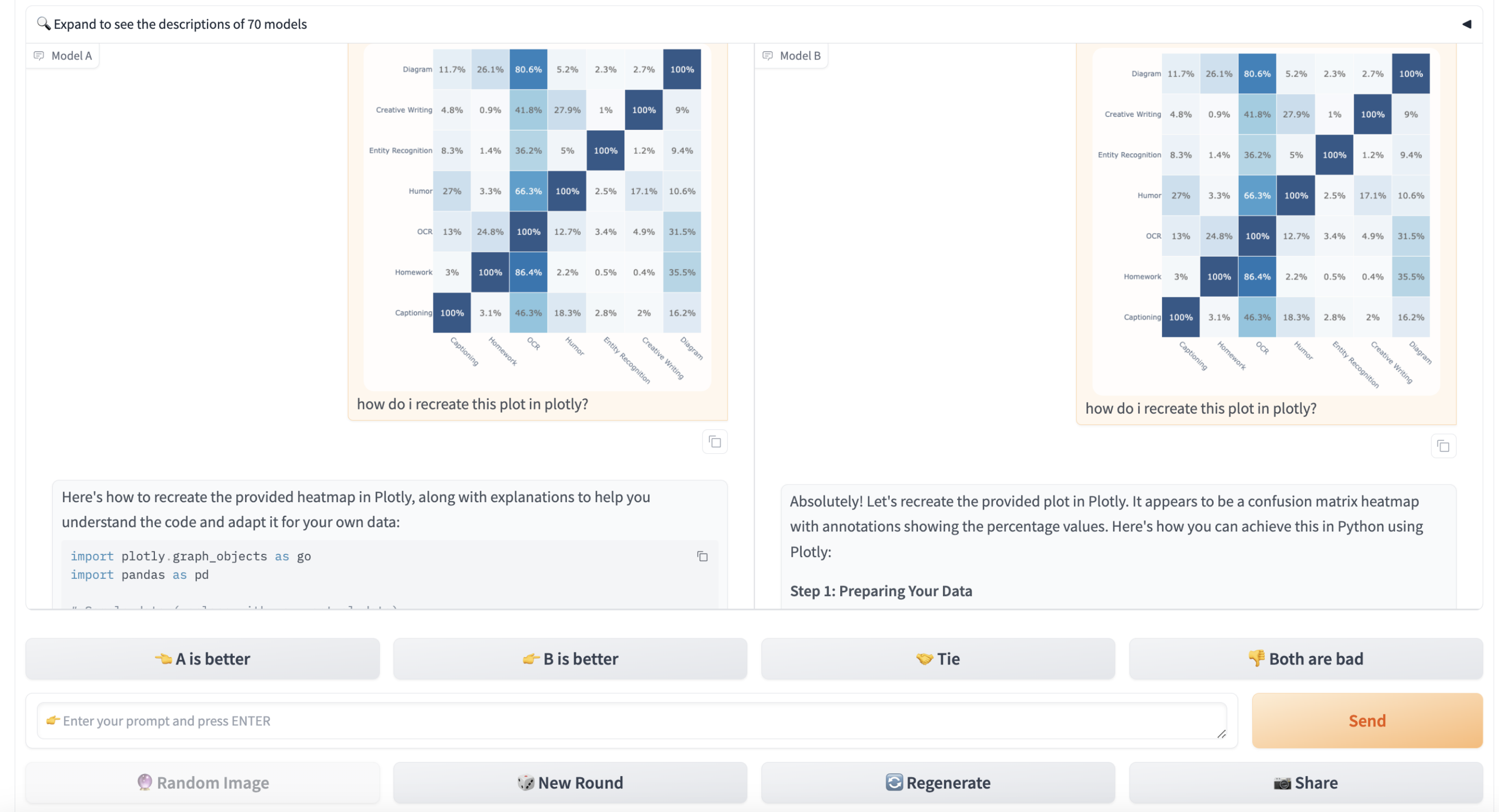}
    \caption{\textbf{Interface for anonymous side-by-side chat.}}
    \label{fig:interface}
\end{figure*}

\begin{figure*}[tbp]
    \centering
    \includegraphics[width=\linewidth]{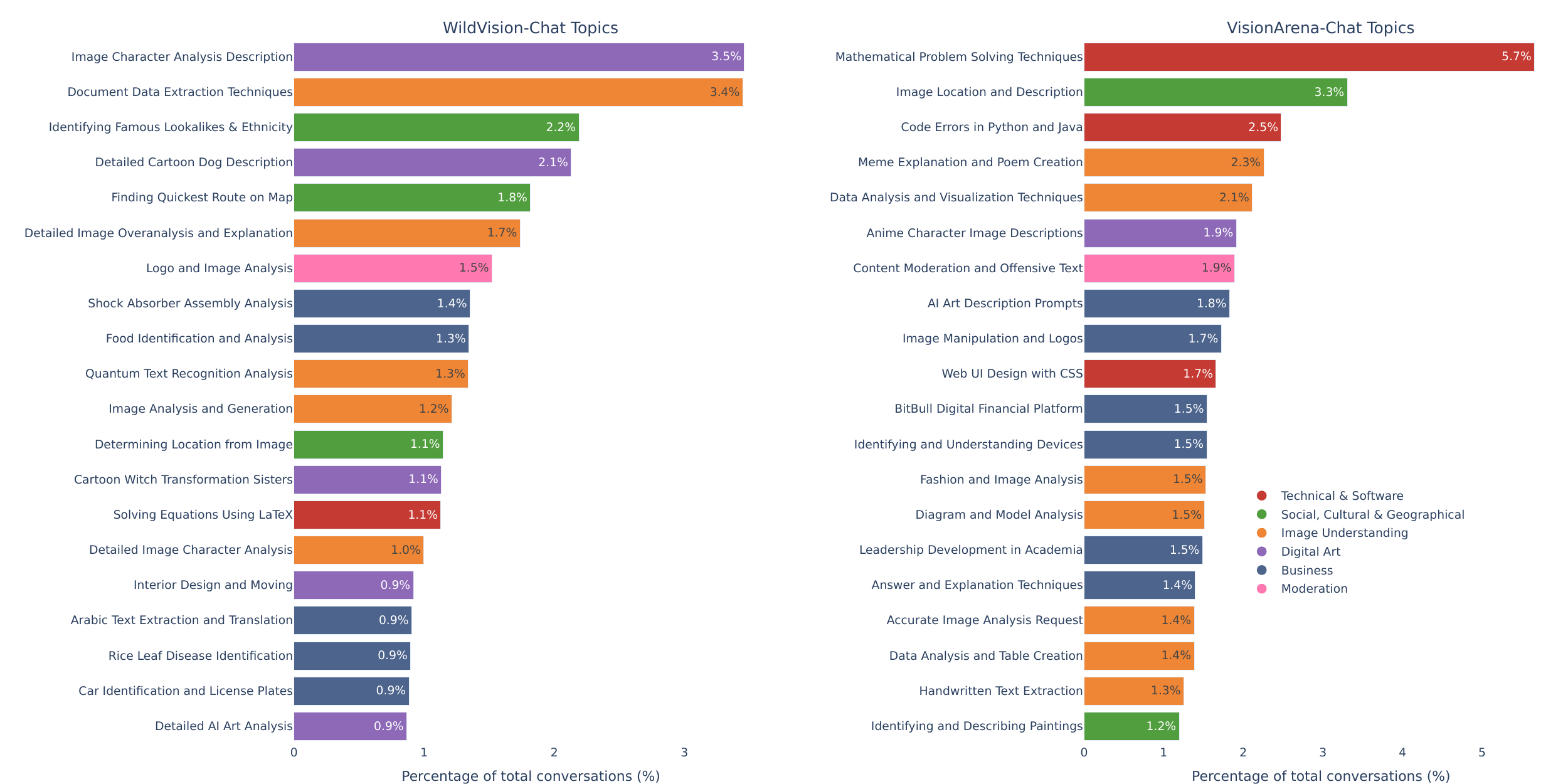}
    \caption{\textbf{Top 20 topic clusters of VisionArena-Chat compared to WildVision-Chat.}  VisionArena-Chat includes more diverse and broad topics especially in the STEM field.}
    \label{fig:topics_20}
    \vspace{2em}
\end{figure*}

\section{Interface Details}
\label{sec:interface_supp}
We implement our interface in Gradio \cite{abid2019gradio}. If a user uploads an image in the first turn, two random VLMs are selected to answer the query. A user can only chat with one image per conversation. As shown in \autoref{fig:interface}, a user can also select a random image which will select from our preset examples. Note that the user still needs to come up with a query, even for these preset images.

\section{Topic Distribution}
\label{sec:topic_distribution}

In \autoref{fig:topics_20} we show the top 20 topic clusters from VisionArena-Chat and WildVision-Chat. The topic clusters are extracted from 50K sampled English conversations from VisionArena-Chat and 37K English conversations (all) from WildVision-Chat. We see that  VisionArena-Chat includes more diverse and broad topics especially in the STEM field. Furthermore, WildVision has very specific clusters like "detailed cartoon dog description", "Shock absorber assembly analysis", "rice lead disease identification". Looking at these clusters we see that they contain a large number of duplicate prompts.

\section{More Data Stats}
\label{sec:data_stats_supp}

\begin{figure}[H]
    \centering
    \includegraphics[width=0.9\linewidth, trim=5 50 30 50, clip]{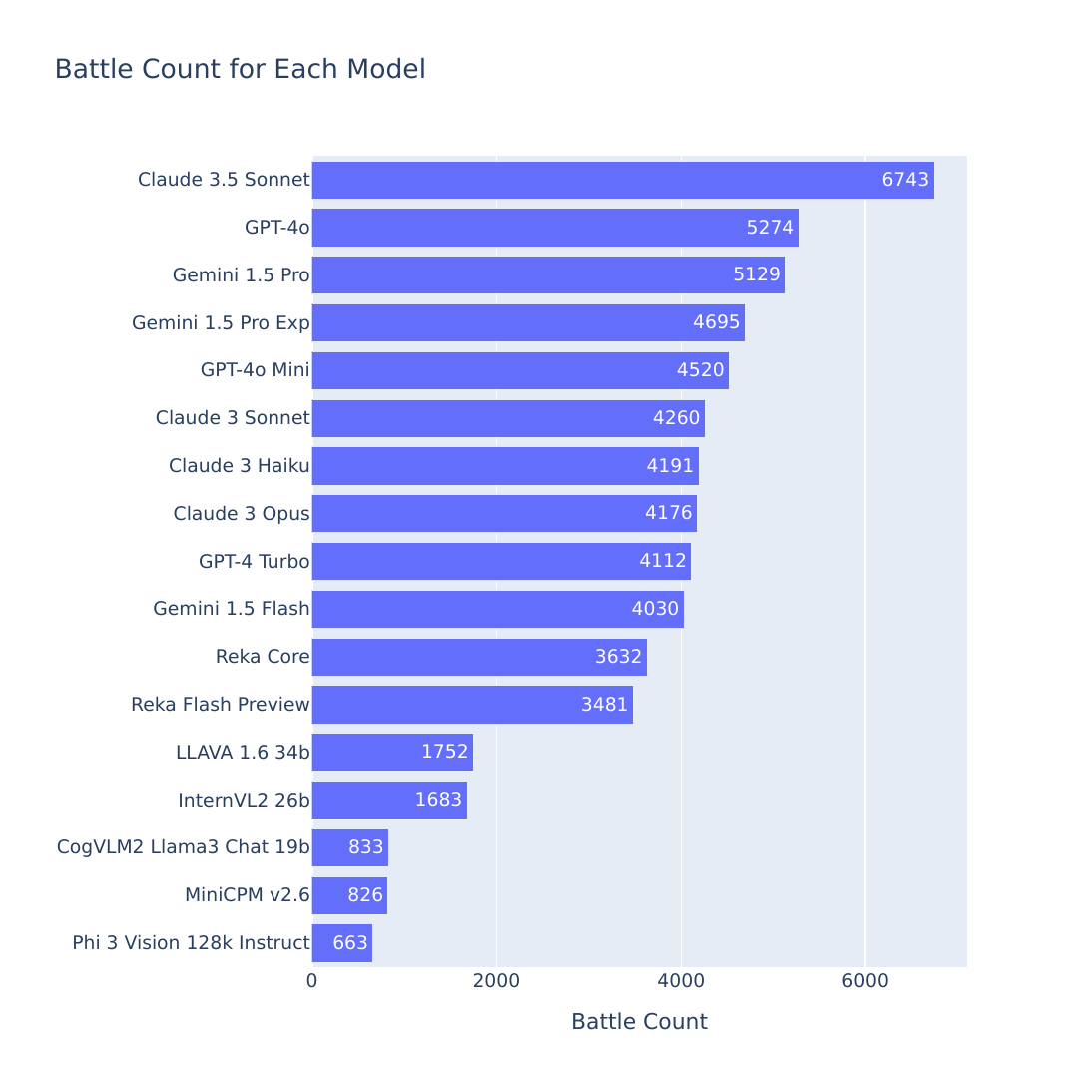}
    \caption{\textbf{VisionArena-Battle counts per model.}}
    \label{fig:model_battle_counts}
\end{figure}

\begin{figure}[H]
    \centering
    \includegraphics[width=0.9\linewidth]{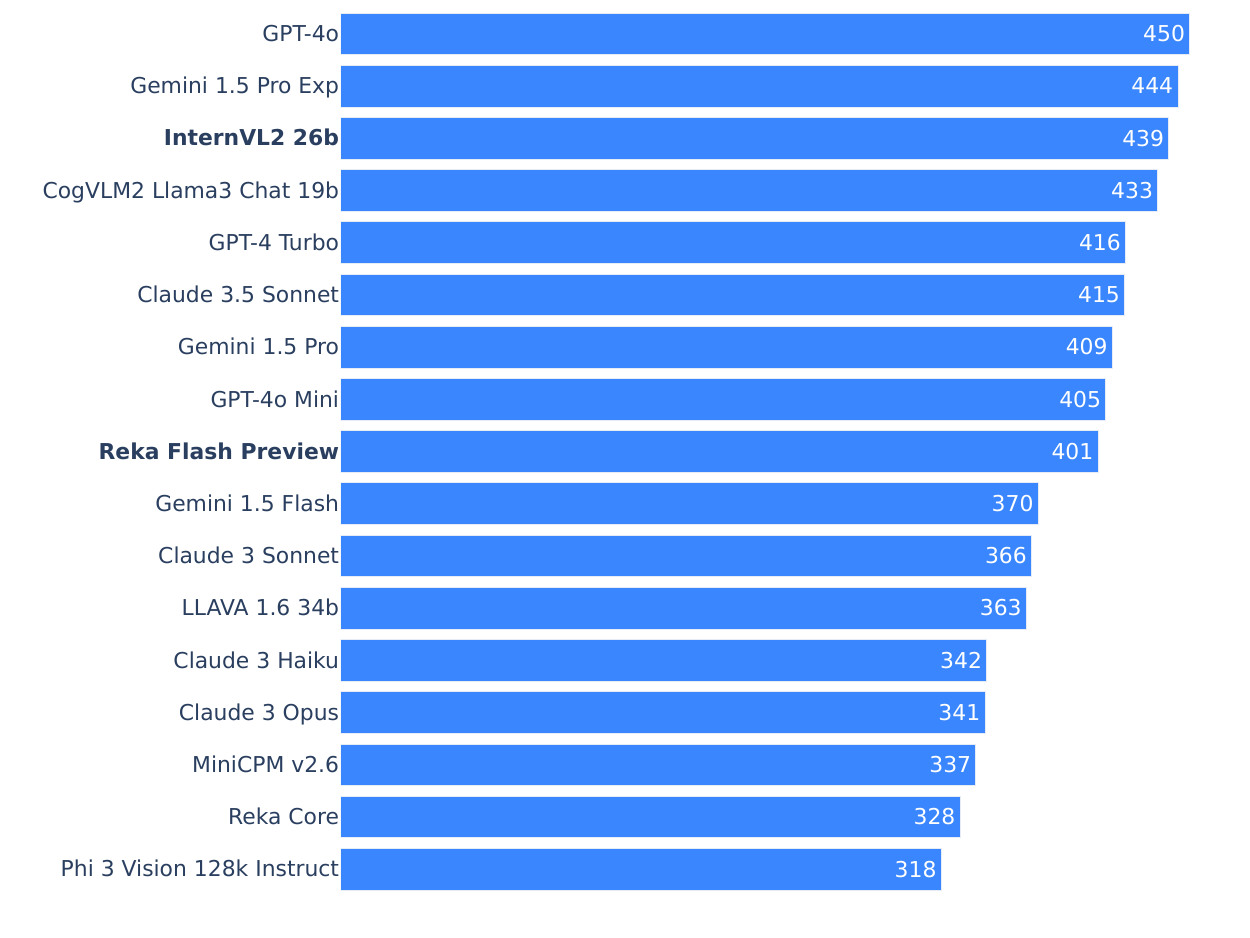}
    \vspace{-1em}
    \caption{\textbf{Model token count in VisionArena-Battle.} Models in bold see a large decrease in rank when style control is applied.}
    \vspace{-2em}
    \label{fig:token_count}
\end{figure}

\begin{figure}[H]
    \centering
    \includegraphics[width=\linewidth, trim=0 50 0 0, clip]{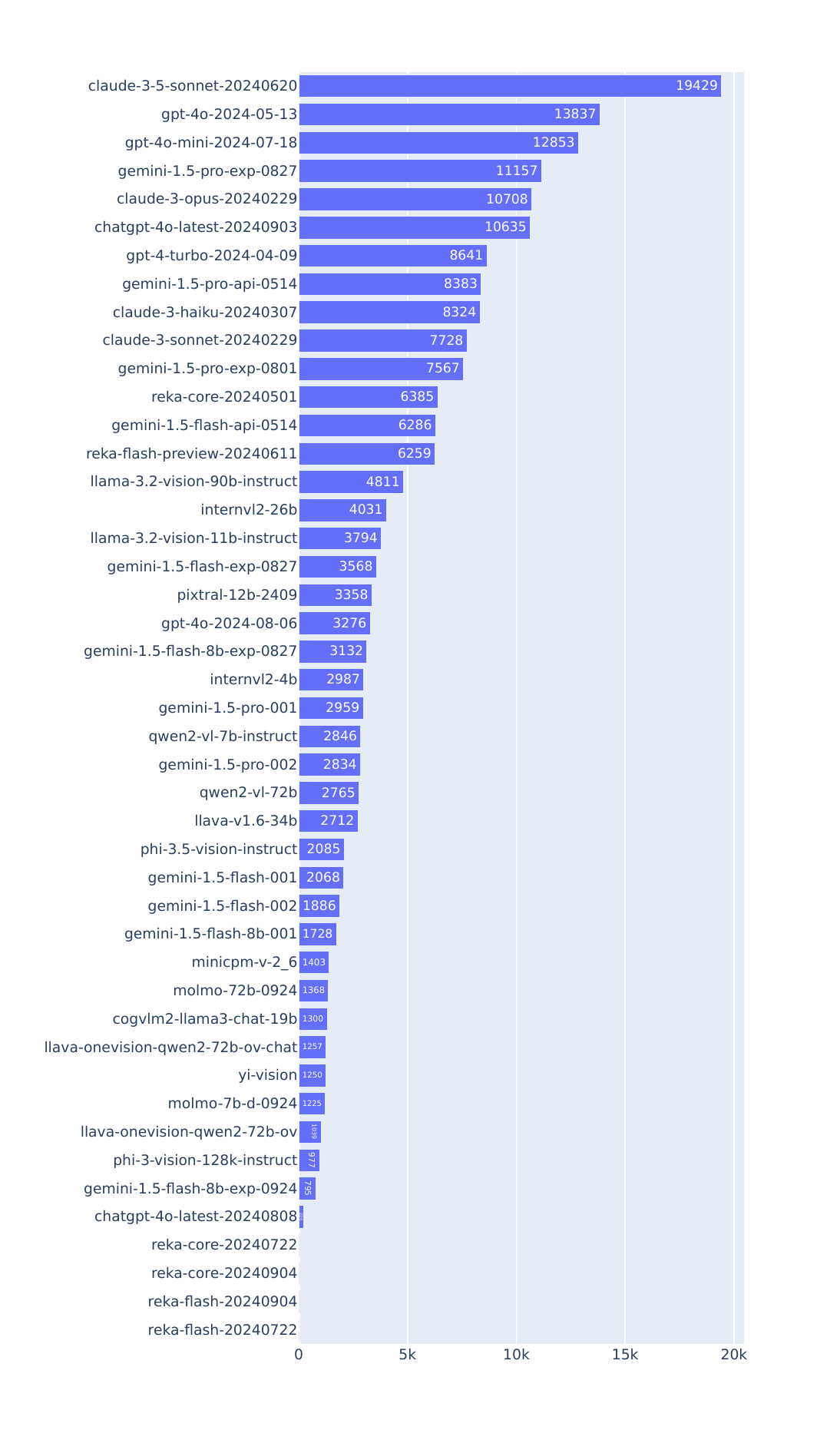}
    \caption{\textbf{VisionArena-Chat counts per model.}}
    \label{fig:direct_model_distribution}
\end{figure}

\begin{figure}[H]
    \centering
    \includegraphics[width=\linewidth, trim=0 40 30 50, clip]{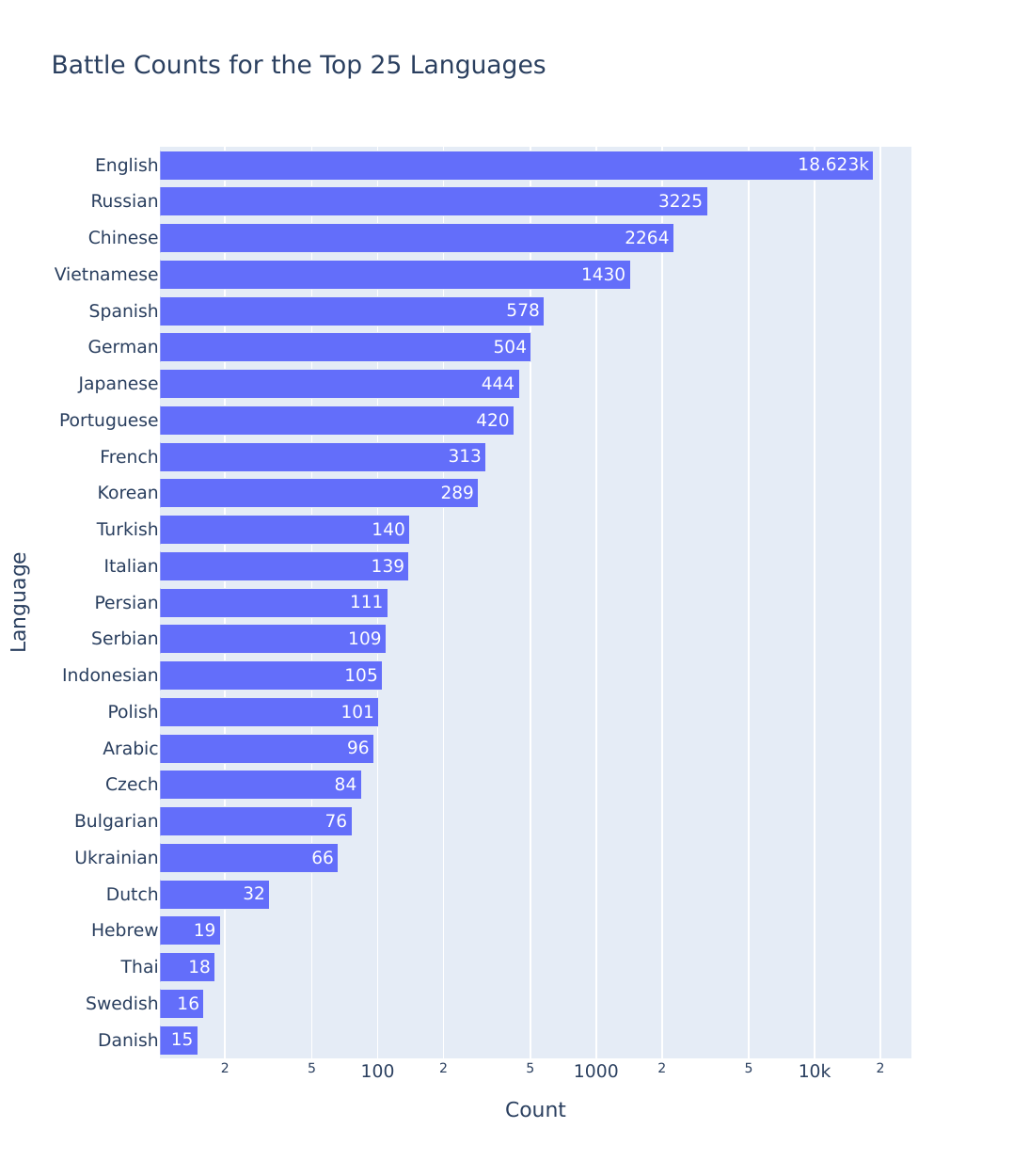}
    \caption{\textbf{VisionArena-Battle counts for the top 25 languages.}}
     \vspace{-1em}
    \label{fig:lang_battle_counts}
\end{figure}

\begin{figure}[H]
    \centering
    \includegraphics[width=\linewidth, trim=10 40 30 80, clip]{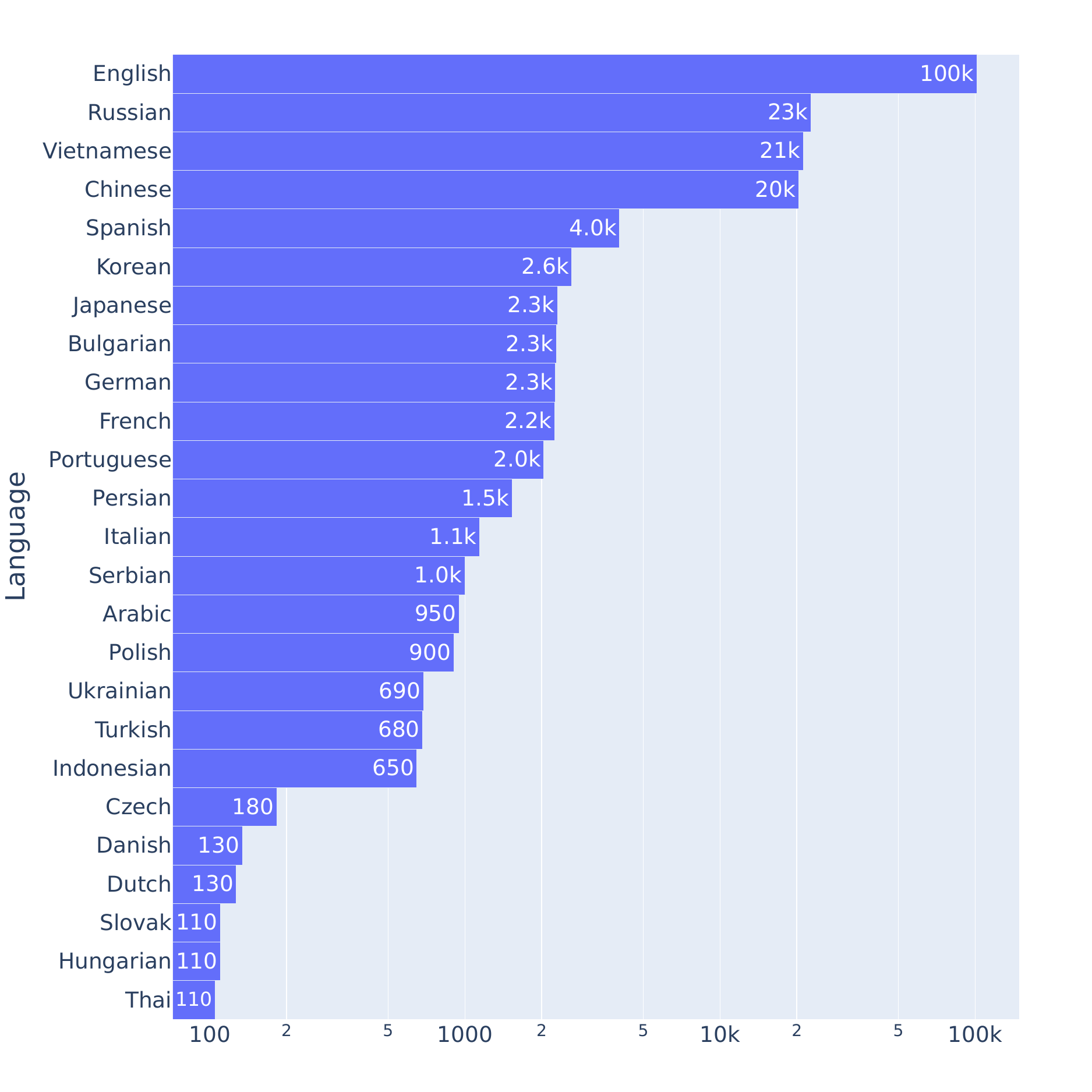}
    \caption{\textbf{VisionArena-Chat counts for the top 25 languages.}}
    \vspace{-1em}
    \label{fig:direct_lang_distribution}
\end{figure}

\begin{figure}[H]
    \centering
    \includegraphics[width=\linewidth, trim=10 40 30 80, clip]{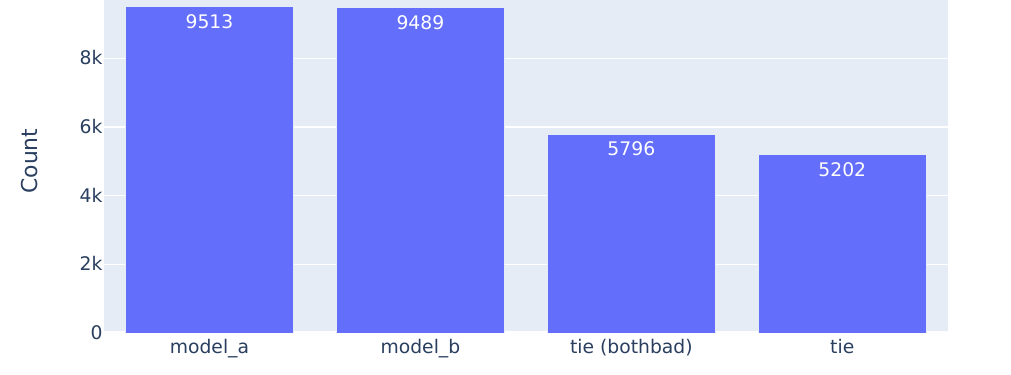}
    \caption{\textbf{Battle Outcome Counts.}}
    \label{fig:battle_outcome_counts}
\end{figure}

\begin{figure}[tbp]
    \centering
    \includegraphics[width=\linewidth, trim=10 40 30 80, clip]{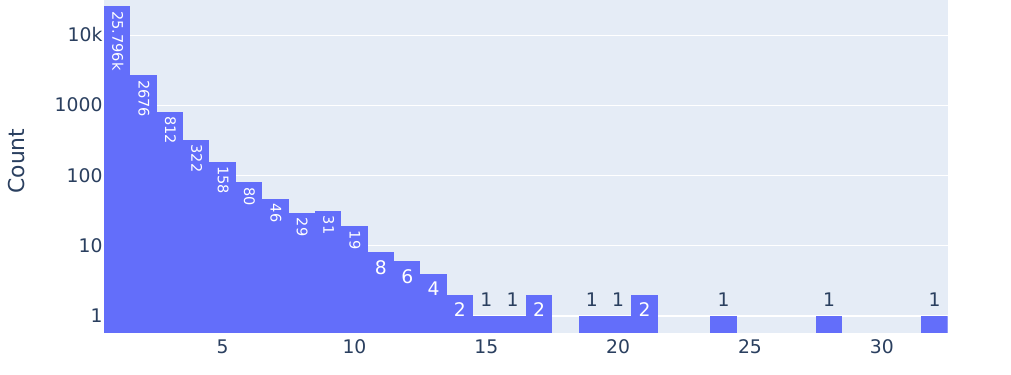}
    \caption{\textbf{VisionArena-Battle Conversation Turn Distribution}}
    \label{fig:turn_distribution}
\end{figure}

\begin{figure}[tbp]
    \centering
    \includegraphics[width=\linewidth, trim=0 40 30 40, clip]{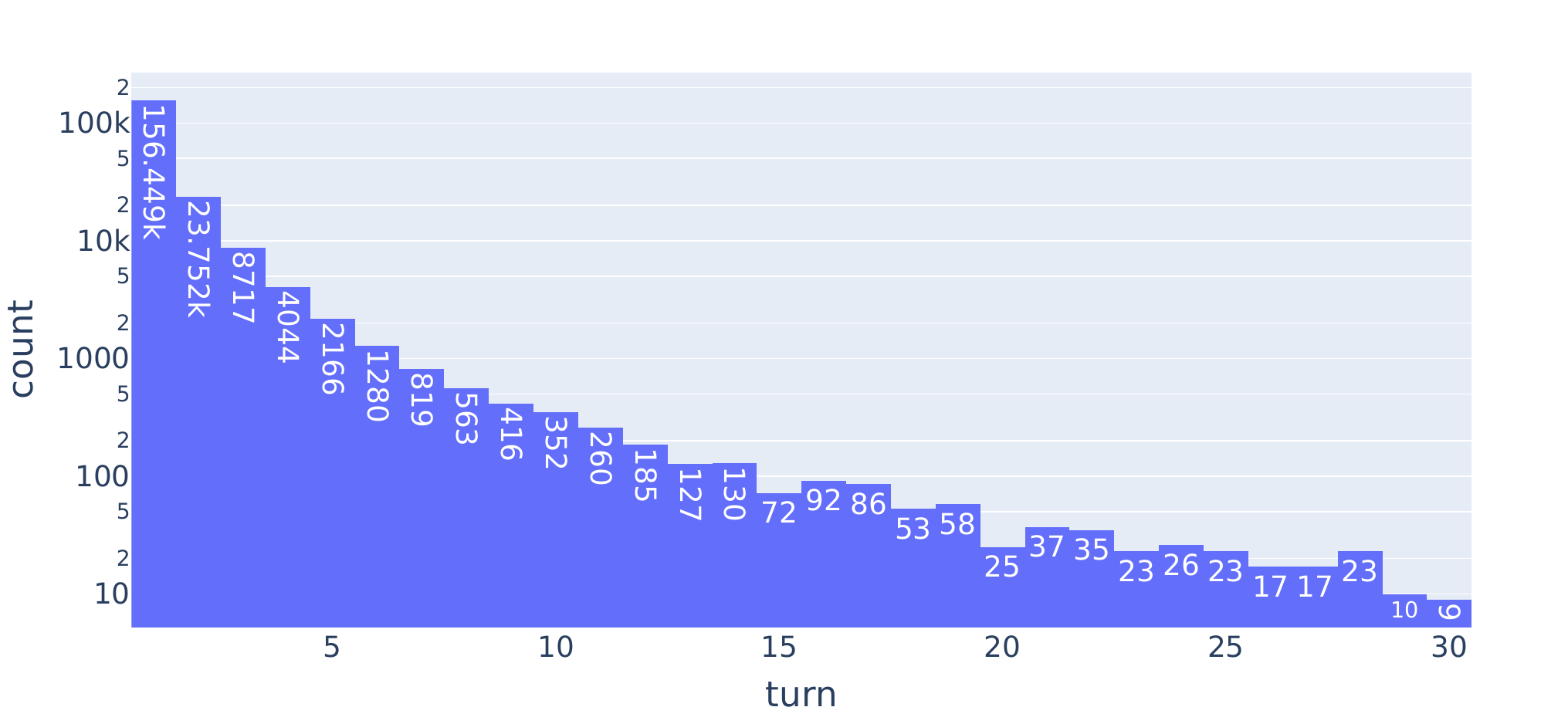}
    \caption{\textbf{VisionArena-Chat Conversation Turn Distribution}}
    \label{fig:direct_turn_distribution}
\end{figure}

\begin{figure}[tbp]
    \centering
    \includegraphics[width=\linewidth, trim=10 10 5 5, clip]{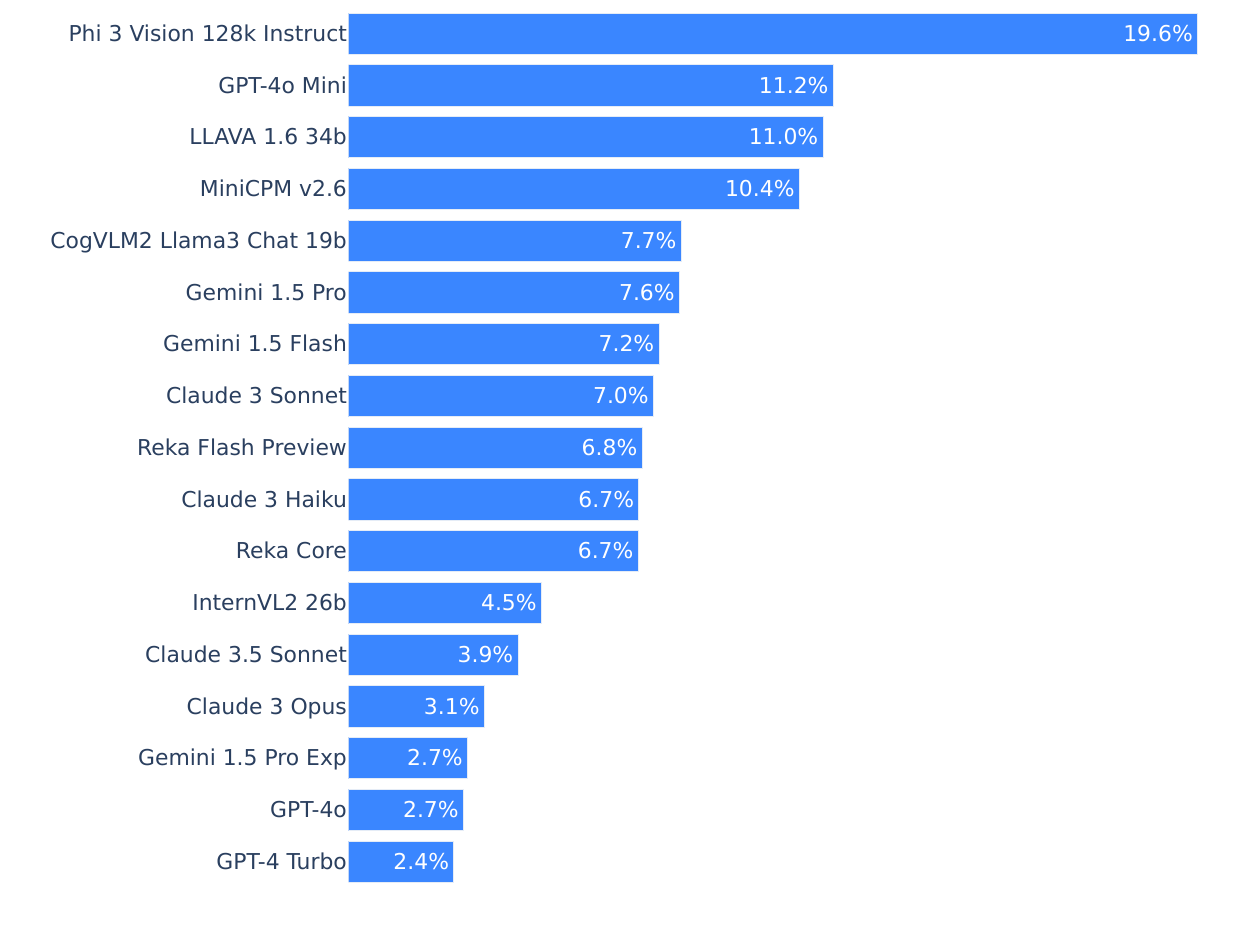}
    \caption{\textbf{Proportion of Refusals per model.}}
    \label{fig:refusals}
\end{figure}

\begin{figure*}[h]
    \begin{subfigure}{0.5\textwidth}
    \includegraphics[width=0.9\linewidth, trim=0 0 0 50, clip]{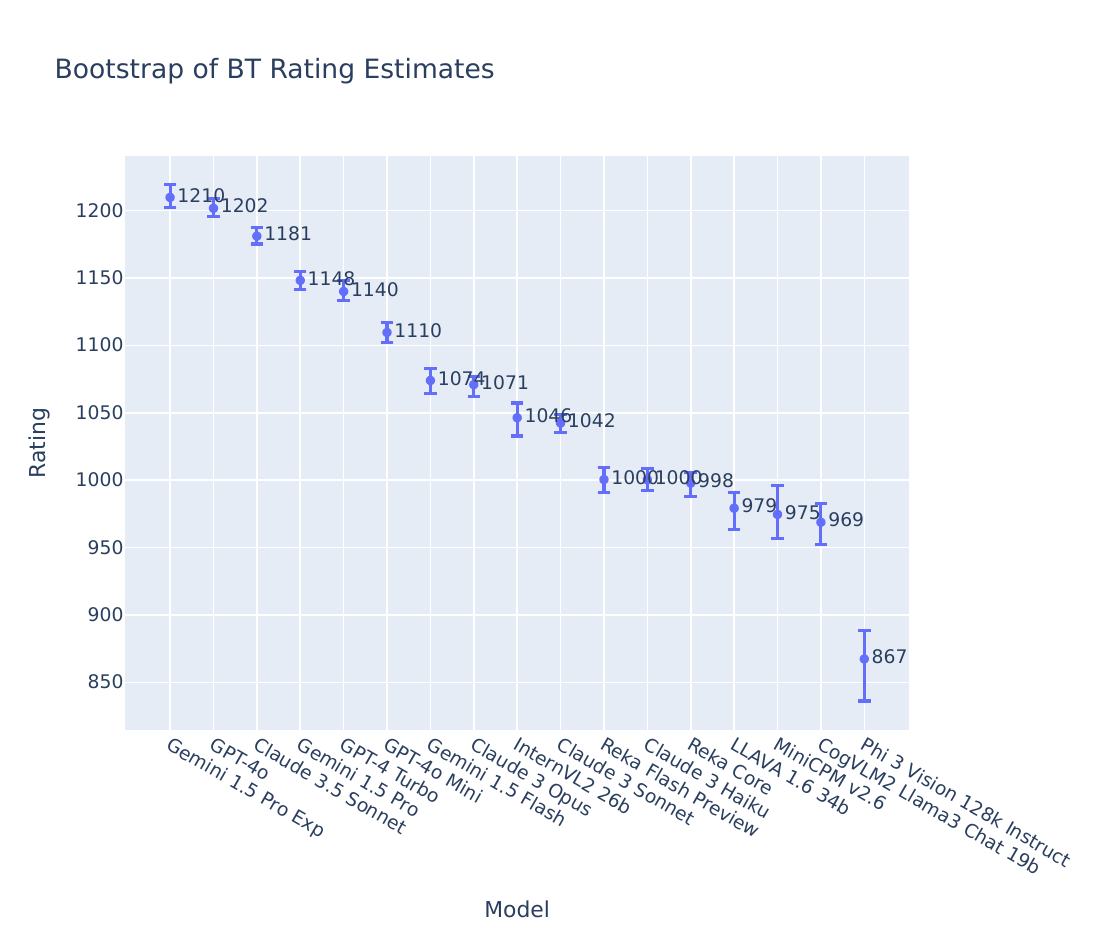}
    \caption{Bootstrap ELO Ratings}
    \label{fig:entity_recognition_bootstrap_elo_ratings}
    \end{subfigure}
    \begin{subfigure}{0.5\textwidth}
    \includegraphics[width=0.9\linewidth, trim=0 0 0 50, clip]{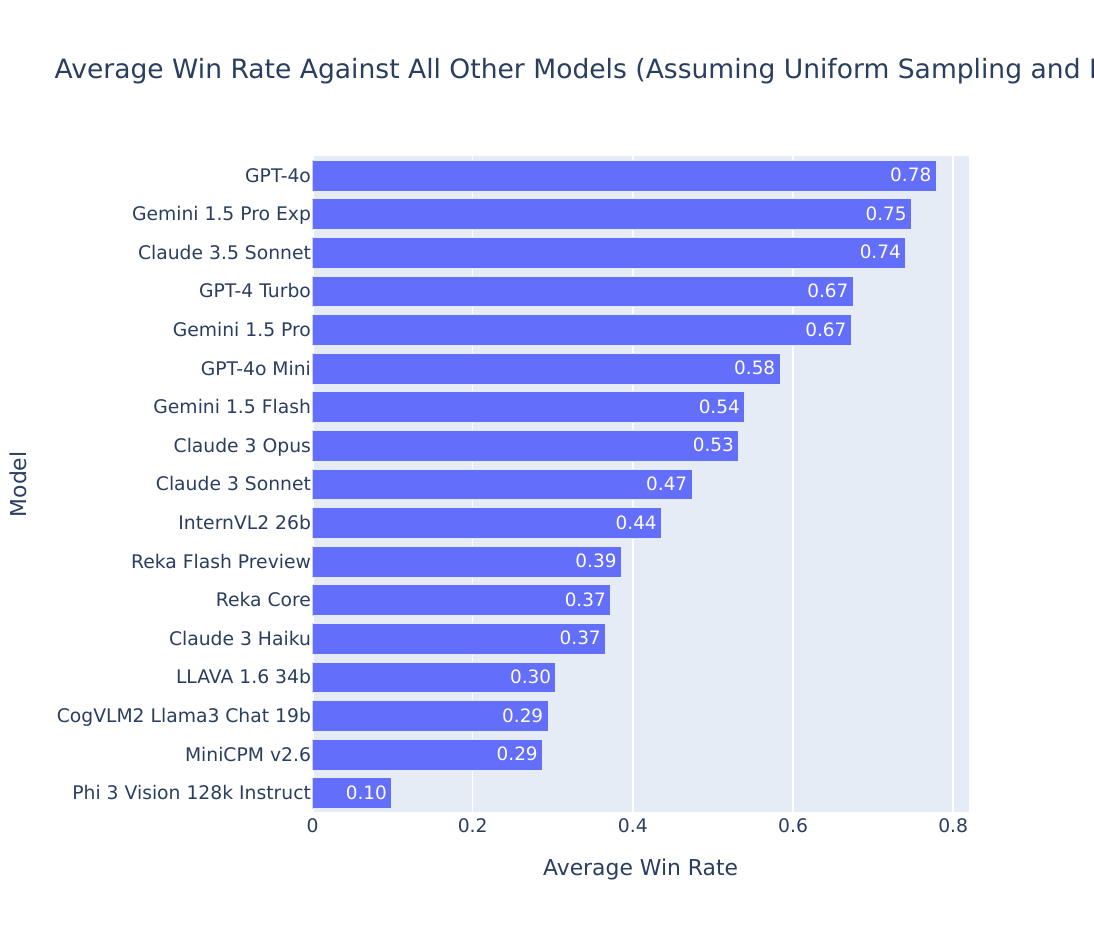}
    \caption{Average Win Rate}
    \label{fig:entity_recognition_average_win_rates}
    \end{subfigure}
    \begin{subfigure}{0.5\textwidth}
    \includegraphics[width=0.8\linewidth, height=6cm]{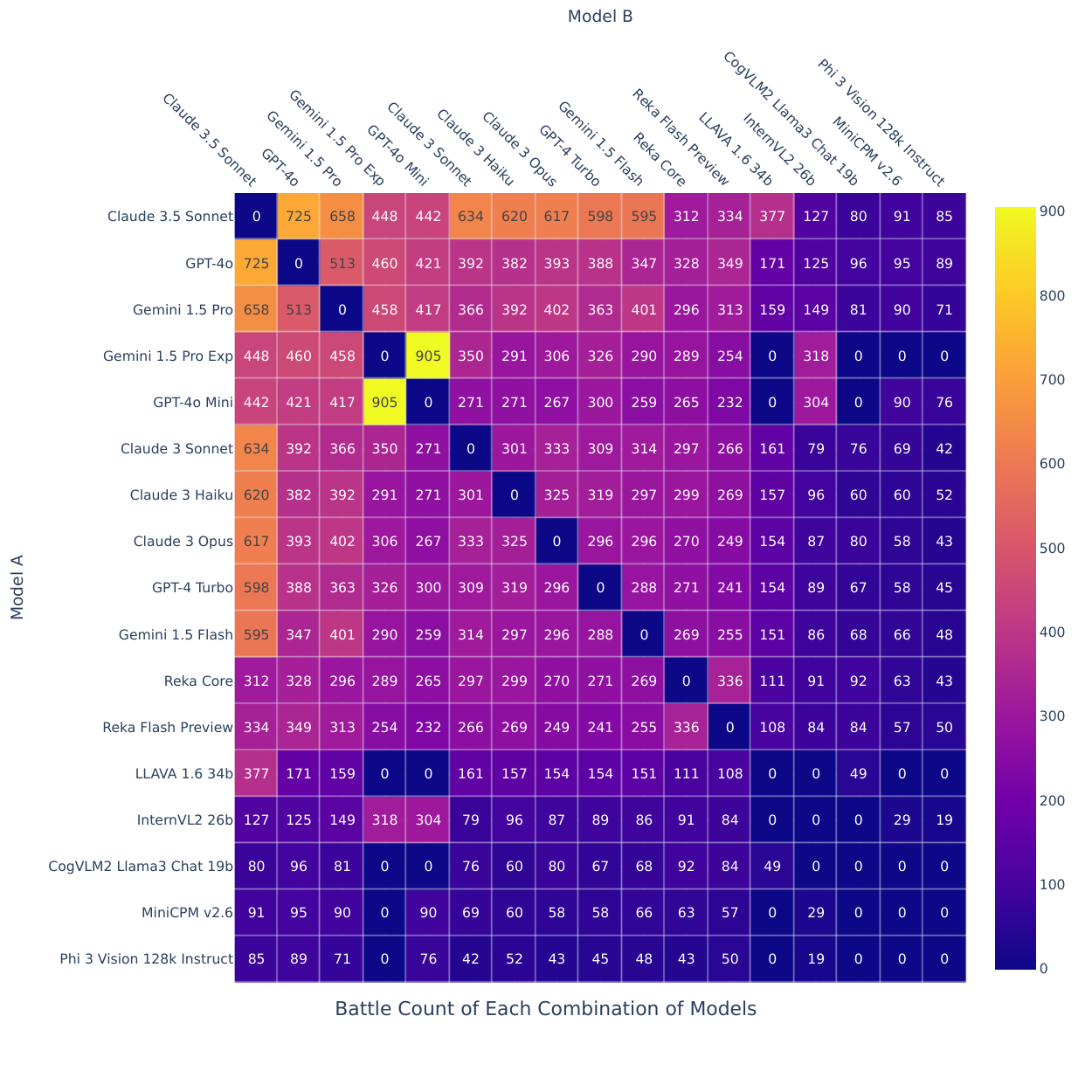}
    \caption{Battle Counts}
    \label{fig:entity_recognition_battle_counts}
    \end{subfigure}
    \begin{subfigure}{0.5\textwidth}
    \includegraphics[width=0.8\linewidth, height=6cm]{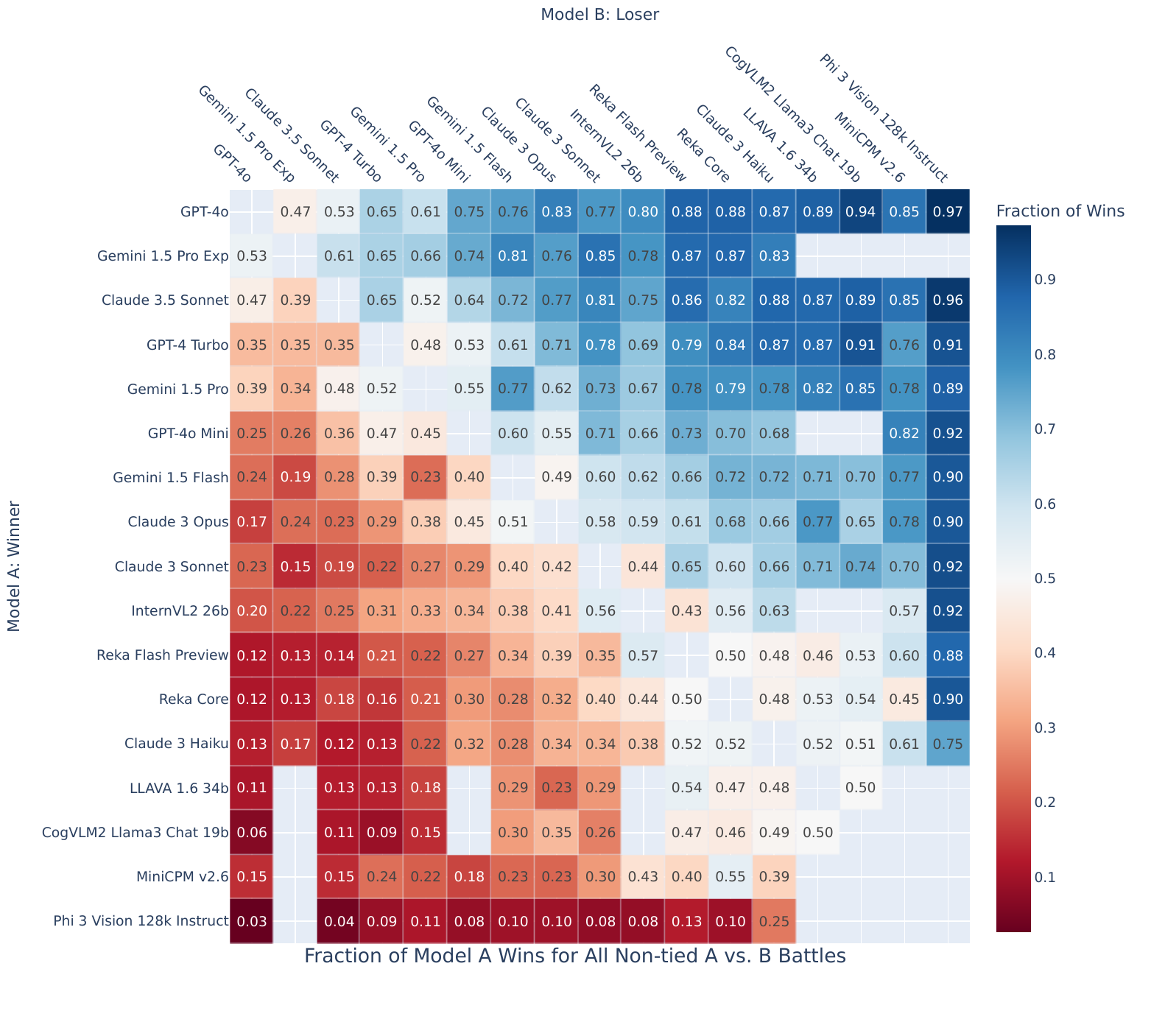}
    \caption{Win Fractions}
    \label{fig:entity_recognition_win_fractions}
    \end{subfigure}
    \caption{\textbf{VisionArena-Battle Model Ranking Results.}}
    \label{fig:win_stats}
\end{figure*}

\begin{figure}[h]
    \centering
    \includegraphics[width=\linewidth]{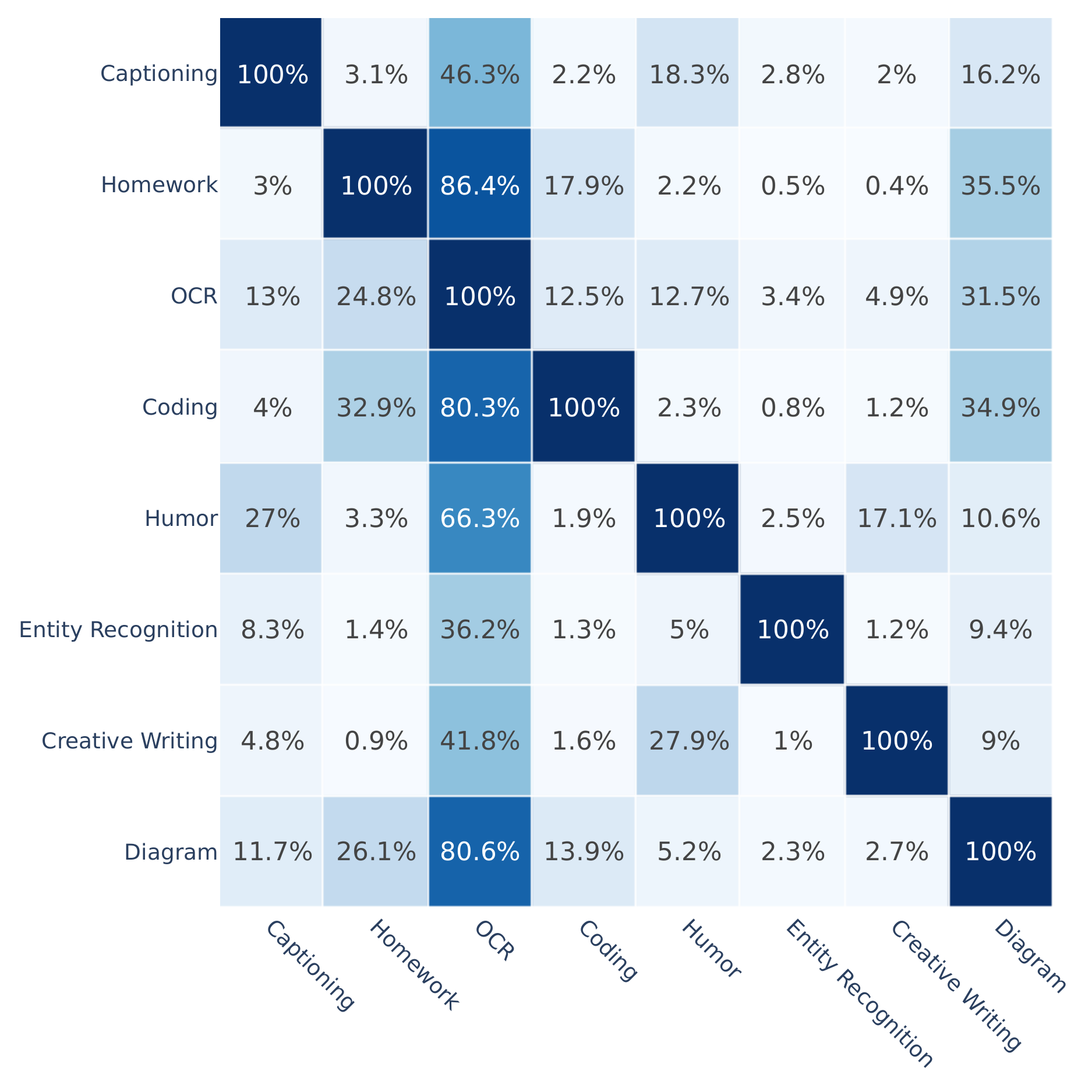}
    \caption{\textbf{VisionArena-Battle category overlap.}}
    \vspace{-1em}
    \label{fig:category_overlap}
\end{figure}

We provide further information on language distribution (\autoref{fig:lang_battle_counts}, \autoref{fig:direct_lang_distribution}), battle counts (\autoref{fig:model_battle_counts}, \autoref{fig:direct_model_distribution}),  token count (\autoref{fig:token_count}), turn distribution (\autoref{fig:turn_distribution}, \autoref{fig:direct_turn_distribution}), proportion of refusals (\autoref{fig:refusals}), battle outcome counts (\autoref{fig:battle_outcome_counts}), win rates (\autoref{fig:win_stats}), and category overlap (\autoref{fig:category_overlap}).

\begin{figure*}
    \centering
    \vspace{4em}
    \includegraphics[width=\linewidth]{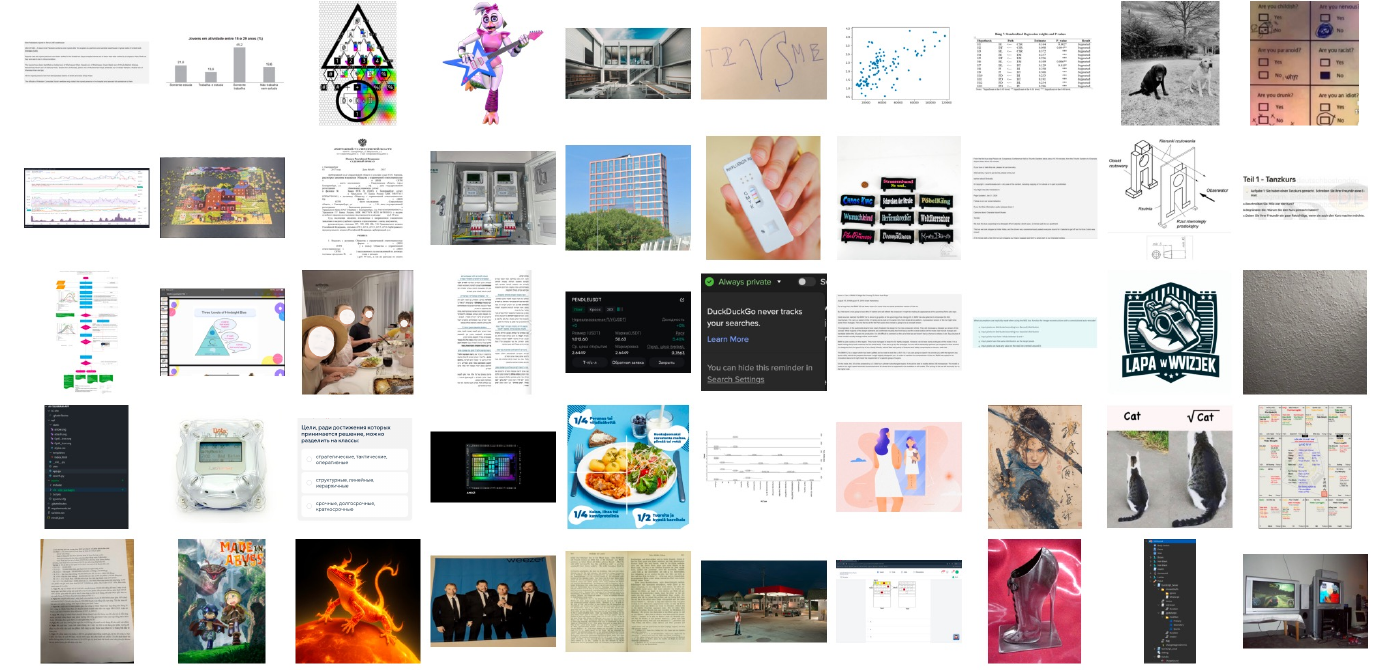}
    \caption{\textbf{Random Samples from VisionArena-Battle}}
    \label{fig:random_samples}
    \vspace{2em}
\end{figure*}

\begin{figure*}
    \centering
    \includegraphics[width=\linewidth]{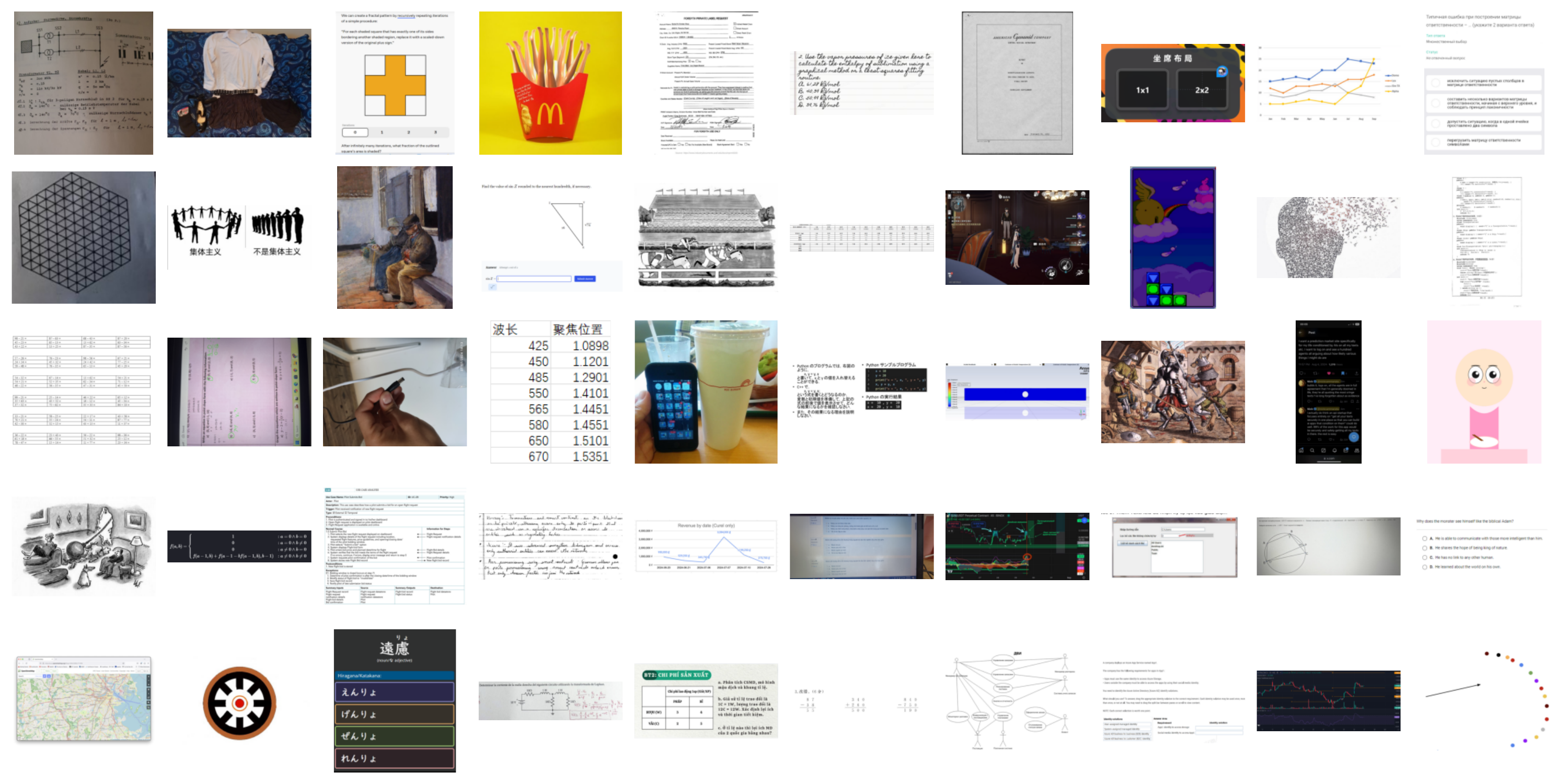}
    \caption{\textbf{Random Samples from VisionArena-Chat}}
    \label{fig:random_dc_samples}
    \vspace{4em}
\end{figure*}

\section{Category Details}
\label{sec:category_supp}

Below are the system prompts used to classify user prompts into the categories described in Section~\ref{sec:analysis}. We find classifications are more accurate for certain categories when using only the prompt or only the image, as indicated in the prompt titles. We use Gemini 1.5 Flash for classification and show in \autoref{tab:category_compare_to_sota} that our classifications achieve high agreement to using SOTA models as category labelers.  

\vspace{5em}

\begin{prompt}{OCR System Prompt (image + prompt)}
You are tasked with determining if a given VQA question is an optical character recognition (OCR) question. An OCR question requires reading and understanding text in the image to answer. If there is some amount of text in the image and the question requires reading the text in any capacity it should be classified as Optical Character Recognition.

\vspace{0.5em}

Output your verdict in the following format:

"$<$decision$>$[yes/no]$<$decision$>$". Do NOT explain.
\end{prompt}

\begin{prompt}{Refusal System Prompt (responses only)}
You are tasked with determining if any of the given model responses are a refusal to answer. A refusal occurs when the model explicitly declines to answer or complete a task, due to reasons like safety, moderation, or model limitations (e.g. the user asks the model to search the web but it does not have that capability). You will be given the responses of 2 models, A and B and you are to determine if A refuses to answer, B refuses, both refuse, or neither refuse. 

\vspace{0.5em}

Output your verdict in the following format:

"$<$decision$>$[yes/no]$<$decision$>$". Do NOT explain.
\end{prompt}

\begin{prompt}{Captioning System Prompt (prompt only)}
You are tasked with determining if a given VQA question is a captioning question. A captioning question asks for a general, overall description of the entire image. It must be a single, open-ended query that does NOT ask about particular objects, people, or parts of the image, nor require interpretation beyond a broad description of what is visually present. Examples include 'What is happening in this image?', 'Describe this picture.', 'Explain', etc. An example of a non-captioning question is 'Describe what is funny in this picture.' because it asks for a specific interpretation of the image content. 

\vspace{0.5em}

Output your verdict in the following format:

$<$decision$>$[yes/no]$<$decision$>$. Do NOT explain.
\end{prompt}

\vspace{7em}

\begin{prompt}{Homework System Prompt (image only)}
You are tasked with determining if the given image contains a homework or exam question. A homework or exam question typically contains text with a well-defined question or task which asks for a solution. In addition, many homework and exam questions contain multiple choice, equations, and question numbers. You may also see text referring to showing your work or providing justification. Note that documents such as resumes, business cards, records, or personal notes are NOT considered homework or exam questions; homework and exam questions explicitly ask for a solution or explanation.

\vspace{0.5em}

Output your verdict in the following format:

"$<$decision$>$[yes/no]$<$decision$>$". Do NOT explain.
\end{prompt}

\begin{table*}[h]
\centering
\begin{tabular}{lcllll}
\toprule
\textbf{Category} & \textbf{Uses Image} & \textbf{Labeler} & \textbf{Accuracy} & \textbf{Precision} & \textbf{Recall} \\
\midrule
Homework & Yes & gemini-1.5-pro-exp-0827 & 0.987 & 0.929 & 0.967 \\
Captioning & No & claude-3-5-sonnet-20240620 & 0.967 & 0.938 & 0.934 \\
Humor & Yes & gemini-1.5-pro-exp-0827 & 0.925 & 0.788 & 0.636 \\
OCR & Yes & gemini-1.5-pro-exp-0827 & 0.818 & 0.954 & 0.769 \\
Entity Recognition & No & claude-3-5-sonnet-20240620 & 0.952 & 0.728 & 0.830 \\
Creative Writing & No & claude-3-5-sonnet-20240620 & 0.964 & 0.680 & 0.810 \\
Diagram & Yes & gemini-1.5-pro-exp-0827 & 0.961 & 0.858 & 0.953 \\
\bottomrule
\end{tabular}
\caption{\textbf{Comparing Gemini-1.5-Flash question categorization to larger models.} Gemini-1.5-Flash is evaluated against SOTA models on 1000 samples from VisionArena-Chat, using Gemini-1.5-Pro for image-based prompts and Claude-3.5-Sonnet for text-based prompts. Gemini-1.5-Flash achieves high agreement with SOTA models for category classification.}
\label{tab:category_compare_to_sota}
\end{table*}

\begin{prompt}{Humor Systems Prompt (image + prompt)}
You are tasked with determining if a given VQA question is a humor question. A humor question asks for a humorous or funny response based on the image or asks to understand what is funny about an image. This includes questions that ask to explain an image which is humorous, such as memes.

\vspace{0.5em}

Output your verdict in the following format:

"$<$decision$>$[yes/no]$<$decision$>$". Do NOT explain.
\end{prompt}

\begin{prompt}{Entity Recognition System Prompt (prompt only)}
You are tasked with determining if a given VQA question is an entity recognition question. An entity recognition question asks for the identification of specific objects or people in the image. This does NOT include questions that ask for a general description of the image, questions that only ask for object counts, or questions that only require reading text in the image.

\vspace{0.5em}

Output your verdict in the following format:

$<$decision$>$[yes/no]$<$decision$>$. Do NOT explain.
\end{prompt}

\begin{prompt}{Diagram System Prompt (image only)}
You are tasked with determining whether the given image contains a chart, diagram, or figure. Carefully examine the user prompt and consider the following aspects:

\vspace{0.5em}

1. Does the image contain visual elements such as graphs, flowcharts, tables, method figures, chemical structures, or other visual representations of data or concepts?

2. Does the prompt require interpreting or analyzing the flow of information, relationships between elements, or the structure of the visual representation in the image?

3. Does the prompt require spatial reasoning and understanding the layout or structure of the visual elements? 

4. Does the image contain only text, tables, handwriting, or photographs without any visual representations of data or concepts? If so, it is NOT considered a chart or diagram.

\vspace{0.5em}

Output your verdict in the following format:

"$<$decision$>$[yes/no]$<$decision$>$". Do NOT explain.
\end{prompt}

\begin{prompt}{Creative Writing System Prompt (prompt only)}
You are tasked with determining whether a given VQA user prompt is asking for creative writing. Creative writing is defined as any form of writing that goes beyond standard professional, journalistic, academic, or technical literature. It typically involves imagination, originality, and expression of thoughts and emotions. Prompts which only ask to caption the image without any other requests do NOT count as creative writing. Creative writing can include, but is not limited to, the following formats:

- Fiction (e.g., short stories, novels),

- Poetry (e.g., sonnets, free verse),

- Dramatic writing (e.g., screenplays, monologues, scripts),

- Personal essays (focusing on subjective experiences or narrative storytelling),

- Songs and lyrics

\vspace{0.5em}

Carefully analyze the user prompt and consider whether it primarily requires creative writing. Think about the following aspects:

1. Does the prompt ask for fictional content, speculative scenarios, or the use of imagination to construct narratives?

2. Does it encourage the expression of thoughts, emotions, or personal experiences beyond mere factual reporting or analysis?

3. Is it asking for writing in a specific creative format (e.g., story, poem, script, etc)?

4. Is the primary purpose of the prompt to foster creative expression or originality rather than information delivery, technical documentation, or analytical reasoning?

5. Does the prompt request stylistic or rhetorical elements often associated with creative writing, such as metaphor, imagery, dialogue, etc?

6. Does the prompt expect a response in natural language (e.g., sentences, paragraphs) rather than visual, mathematical, or non-linguistic output?

\vspace{0.5em}

Output your verdict in the following format:

"$<$decision$>$[yes/no]$<$decision$>$". Do NOT explain.
\end{prompt}

\section{Contamination with Existing Benchmarks}
\label{sec:contamination_supp}

To ensure that our results from~\autoref{sec:finetuning} are not due to training on questions from the test sets, we investigate the rate of benchmark contamination in \dataset{}-Chat. Using OpenAI’s text-embedding-small embeddings, we compute the cosine similarity between each VisionArena-Chat question and all benchmark questions, selecting the nearest neighbor with the highest similarity score. We then count the number of cases where this similarity is $\geq$ 0.8, indicating minor rephrasings of the same question. \autoref{tab:contamination} shows that less than 2\% of benchmark questions are seen on \dataset{}-Chat.

\begin{table}[h!]
% \vspace{-1em}
\centering
\resizebox{\columnwidth}{!}{
\begin{tabular}{lcccc}
\toprule
\textbf{Dataset} & \# Matches & \% dataset & \% \dataset{}-Chat \\
\midrule
MMMU & 47 & 0.4\% & 0.02\%\\
MME & 0 & 0.0\% & 0.0\%\\
HallusionBench & 0 & 0.0\% & 0.00\% \\
MMMU Pro & 23 & 1.3\% & 0.01\% \\
\bottomrule
\end{tabular}
}
\vspace{-1em}
\caption{\textbf{Proportion of benchmark data in \dataset{}-Chat.}}
\label{tab:contamination}
\end{table}

% \resizebox{0.5\textwidth}{!}{\begin{tabular}{lccc}

% \toprule
%                       \textbf{Model} &  \textbf{Ranking (No Modification)} &  \textbf{Ranking (Hallucination Control)} & \textbf{Ranking (Specificity Control)} \\
% \midrule
%     gemini-1.5-pro-exp-0801 &        1 & 1 & 2\\
%           gpt-4o-2024-05-13 &        2 & 2 & 1\\
%  claude-3-5-sonnet-20240620 &        3 & 3 & 3\\
%     gemini-1.5-pro-api-0514 &        4 & 4 & 4\\
%      gpt-4-turbo-2024-04-09 &        5 & 5 & 5\\
%      gpt-4o-mini-2024-07-18 &        6 & 6 & 6\\
%   gemini-1.5-flash-api-0514 &        7 & 7 & 7\\
%      claude-3-opus-20240229 &        8 & 8 & 8 \\
%               internvl2-26b &        9 & 10 & 10 \\
%    claude-3-sonnet-20240229 &       10 & 9 & 9 \\
% reka-flash-preview-20240611 &       11 & 14 &  13\\
%     claude-3-haiku-20240307 &       12 & 11 & 11 \\
%          reka-core-20240501 &       13 & 12 & 12\\
%              llava-v1.6-34b &       14 &  15 & 15\\
%               minicpm-v-2\_6 &       15 & 13 & 14 \\
%     cogvlm2-llama3-chat-19b &       16 &  16 & 16 \\
%  phi-3-vision-128k-instruct &       17 &  17 & 17 \\
% \bottomrule
% \end{tabular}
% }

\section{VisionArena-Bench}
\label{sec:vision_arena_bench_supp}

\begin{prompt}{VLM-as-a-Judge System Prompt}
\label{prompt:vlm_judge}
Please act as an impartial judge and evaluate the quality of the responses provided by two AI assistants to
the user prompt displayed below. You will be given assistant A’s answer and assistant B’s answer. Your
job is to evaluate which assistant’s answer is better.

% \vspace{0.5em}

Begin your evaluation by generating your own answer to the prompt. You must provide your answers before
judging any answers.

When evaluating the assistants’ answers, compare both assistants’ answers with your answer. You must identify and correct any mistakes or inaccurate information.

% \vspace{0.5em}

Then consider if the assistant’s answers are helpful, relevant, and concise. Helpful means the answer
correctly responds to the prompt or follows the instructions. Note when user prompt has any ambiguity or
more than one interpretation, it is more helpful and appropriate to ask for clarifications or more information
from the user than providing an answer based on assumptions. Relevant means all parts of the response
closely connect or are appropriate to what is being asked. Concise means the response is clear and not
verbose or excessive.

% \vspace{0.5em}

Then consider the creativity and novelty of the assistant’s answers when needed. Finally, identify any
missing important information in the assistants’ answers that would be beneficial to include when responding
to the user prompt.

% \vspace{0.5em}

After providing your explanation, you must output only one of the following choices as your final verdict
with a label:
1. Assistant A is significantly better: $[[A>>B]]$

2. Assistant A is slightly better: $[[A>B]]$

3. Tie, relatively the same: $[[A=B]]$

4. Assistant B is slightly better: $[[B>A]]$

5. Assistant B is significantly better: $[[B>>A]]$

Example output: "My final verdict is tie: $[[A=B]]$".
\end{prompt}

\label{sec:bench}

\begin{table}[H]
\label{table:arena_bench}
\small
\centering
\begin{tabular}{lccc}
\toprule
\textbf{Model} & \textbf{Score} & \textbf{95\% CI} & \textbf{Token \#} \\
\midrule
gpt-4o-2024-05-13 & 67.7 & (-1.7, 1.8) & 316 \\
gemini-1.5-pro-exp-0827 & 66.2 & (-1.8, 1.5) & 329 \\
gemini-1.5-flash-exp-0827 & 60.3 & (-1.9, 1.9) & 367 \\
claude-3.5-sonnet-20240620 & 54.5 & (-2.1, 1.9) & 262 \\
gpt-4-turbo-2024-04-09 & 50.0 & (0.0, 0.0) & 258 \\
gemini-1.5-pro-001 & 45.5 & (-1.8, 2.0) & 261 \\
gpt-4o-mini-2024-07-18 & 40.0 & (-2.3, 1.9) & 224 \\
gemini-1.5-flash-8b-exp-0827 & 30.6 & (-2.3, 1.8) & 354 \\
internvl2-26b & 23.3 & (-2.1, 1.1) & 515 \\
gemini-1.5-flash-001 & 23.0 & (-1.1, 1.6) & 271 \\
claude-3-opus-20240229 & 18.9 & (-1.9, 1.7) & 201 \\
claude-3-sonnet-20240229 & 18.4 & (-1.4, 1.3) & 205 \\
reka-core-20240501 & 15.6 & (-1.3, 1.4) & 252 \\
llama-3.2-11b-vision-instruct & 11.2 & (-1.3, 1.1) & 457 \\
claude-3-haiku-20240307 & 9.6 & (-1.1, 1.0) & 155 \\
internvl2-4b & 6.8 & (-0.9, 0.8) & 421 \\
\bottomrule
\end{tabular}
\caption{VisionArena-Bench leaderboard (baseline: GPT-4-Turbo)}
\end{table}

\begin{table}[H]
\label{table:bench_sonnet}
\small
\centering
\begin{tabular}{lccc}
\toprule
\textbf{Model} & \textbf{Score} & \textbf{95\% CI} & \textbf{Token \#} \\
\midrule
gemini-1.5-pro-exp-0827 & 87.6 & (-1.1, 1.1) & 329 \\
gpt-4o-2024-05-13 & 86.8 & (-1.1, 1.3) & 316 \\
claude-3-5-sonnet-20240620 & 86.3 & (-1.3, 1.2) & 262 \\
gemini-1.5-flash-exp-0827 & 83.5 & (-1.7, 1.1) & 367 \\
gpt-4-turbo-2024-04-09 & 80.7 & (-1.0, 1.6) & 258 \\
gemini-1.5-pro-001 & 75.3 & (-1.7, 1.4) & 261 \\
gpt-4o-mini-2024-07-18 & 73.0 & (-1.3, 1.4) & 224 \\
gemini-1.5-flash-8b-exp-0827 & 64.7 & (-1.5, 2.4) & 354 \\
gemini-1.5-flash-001 & 58.4 & (-1.9, 1.5) & 271 \\
internvl2-26b & 54.2 & (-1.7, 1.7) & 515 \\
claude-3-opus-20240229 & 52.0 & (-2.0, 1.7) & 201 \\
claude-3-sonnet-20240229 & 50.0 & (0.0, 0.0) & 205 \\
reka-core-20240501 & 37.9 & (-1.9, 1.7) & 252 \\
llama-3.2-11b-vision-instruct & 32.6 & (-1.7, 1.8) & 457 \\
claude-3-haiku-20240307 & 30.7 & (-2.3, 1.6) & 155 \\
internvl2-4b & 19.6 & (-1.9, 1.3) & 421 \\
\bottomrule
\end{tabular}
\caption{VisionArena-Bench leaderboard (baseline: claude-3-sonnet-20240229)}
\end{table}

\section{Additional model details}

\autoref{tab:model_name_to_version} shows the mapping from the model names used in Section~\ref{sec:analysis} to the exact model versions. 

\begin{table}[h]
\centering
\begin{tabular}{ll}
\hline
\textbf{Model Version} & \textbf{Model Name} \\
\hline
claude-3-5-sonnet-20240620 & Claude 3.5 Sonnet \\
claude-3-haiku-20240307 & Claude 3 Haiku \\
claude-3-opus-20240229 & Claude 3 Opus \\
claude-3-sonnet-20240229 & Claude 3 Sonnet \\
cogvlm2-llama3-chat-19b & CogVLM2 Llama3 Chat 19b \\
gemini-1.5-flash-api-0514 & Gemini 1.5 Flash \\
gemini-1.5-pro-api-0514 & Gemini 1.5 Pro \\
gemini-1.5-pro-exp-0801 & Gemini 1.5 Pro Exp \\
gpt-4-turbo-2024-04-09 & GPT-4 Turbo \\
gpt-4o-2024-05-13 & GPT-4o \\
gpt-4o-mini-2024-07-18 & GPT-4o Mini \\
internvl2-26b & InternVL2 26b \\
llava-v1.6-34b & LLAVA 1.6 34b \\
minicpm-v-2\_6 & MiniCPM v2.6 \\
phi-3-vision-128k-instruct & Phi 3 Vision 128k Instruct \\
reka-core-20240501 & Reka Core \\
reka-flash-preview-20240611 & Reka Flash Preview \\
\hline
\end{tabular}
\caption{Model Name to exact model version}
\label{tab:model_name_to_version}
\end{table}

\section{Failure Cases}
\label{sec:failures_supp}

\begin{table*}[tbp]
    \centering
    \small
    \resizebox{0.90\textwidth}{!}{%
    \begin{tabular}{cccccccccc}
    \toprule
    Model & K-Pop & Sign & Shapes & Triangles & Meme & Map & Shoes & Chess & ARC \\
    \midrule
    gemini-1.5-pro-exp-0827         & X & X & X & X & X & X & X & X & X \\
    gpt-4o-2024-05-13               & X & O & X & X & X & X & O & X & X \\
    claude-3-5-sonnet-20240620      & X & X & X & X & X & X & X & X & X \\
    claude-3-opus-20240229          & X & X & X & X & X & X & O & X & X \\
    gpt-4-turbo-2024-04-09          & X & X & X & X & X & X & X & X & X \\
    gpt-4o-mini-2024-07-18          & X & X & X & X & X & X & X & X & X \\
    gemini-1.5-pro-001              & X & X & X & X & X & X & X & X & X \\
    gemini-1.5-flash-8b-exp-0827    & X & X & X & X & X & X & X & X & X \\
    gemini-1.5-flash-exp-0827       & X & O & O & O & X & X & X & X & X \\
    internvl2-26b                   & X & X & X & X & X & X & X & X & X \\
    gemini-1.5-flash-001            & X & X & O & X & X & X & X & X & X \\
    claude-3-sonnet-20240229        & X & X & X & X & X & X & O & X & X \\
    llama-3.2-11b-vision-instruct   & X & O & X & O & X & X & X & X & X \\
    claude-3-haiku-20240307         & X & X & X & X & X & X & X & X & X \\
    internvl2-4b                    & X & X & X & X & X & X & X & X & X \\
    \bottomrule
    \end{tabular}%
    }
    \vspace{-0.5em}
    \caption{\textbf{Model performance across several hard tasks.} O indicates that the model solves the problem and X indicates that the model fails to solve the problem. 9 out of 16 models fail all questions.}
    \label{tab:model_failures}
\end{table*}

% \textit{Hard OCR} (\autoref{fig:failure_cases_ocr}). While VLMs are strong at transcribing text from images where the text is easily legible, they struggle when it is perturbed (e.g. rotations, blur), The ability to read these hard texts is crucial for models to perform in the real world. We show two examples where models fail, one where the handwriting is not clear and the second where the text is rotated.

% \textit{Counting} (\autoref{fig:failure_cases_counting}). Counting is an important skill for decision-making across education, organization, and daily use. While humans can count quite easily, we show that VLMs still fail at these tasks. We show two examples where all of the top 3 models fail, one where the models have to count based on both the shape and color and one where the models have to count triangles that may intersect one another. 

% \textit{Reasoning} (\autoref{fig:hard_reasoning_failures}, \autoref{fig:failure_case_reasoning}). Reasoning in LLMs and VLMs is a crucial skill to help users think about and understand difficult problems. While the gap in reasoning is not restricted to just VLMs and is still an unsolved problem in LLMs, we present 5 unique use cases of VLMs and how they fail.

\begin{figure*}[h]
    \centering
    \begin{subfigure}[b]{0.90\linewidth}
        \centering
        \includegraphics[width=\linewidth]{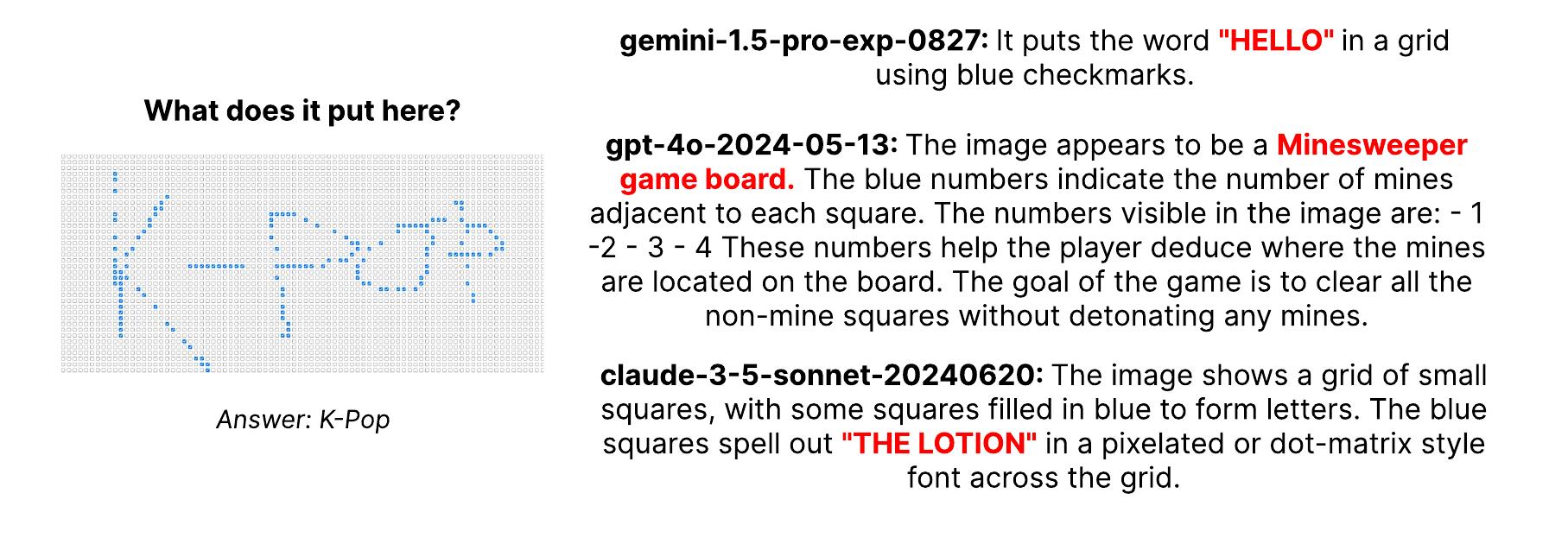}
        \caption{K-Pop Failure Case}
        \label{fig:k_pop}
    \end{subfigure}
    \hfill
    \begin{subfigure}[b]{0.90\linewidth}
        \centering
        \includegraphics[width=0.9\linewidth, trim=5 40 5 50, clip]{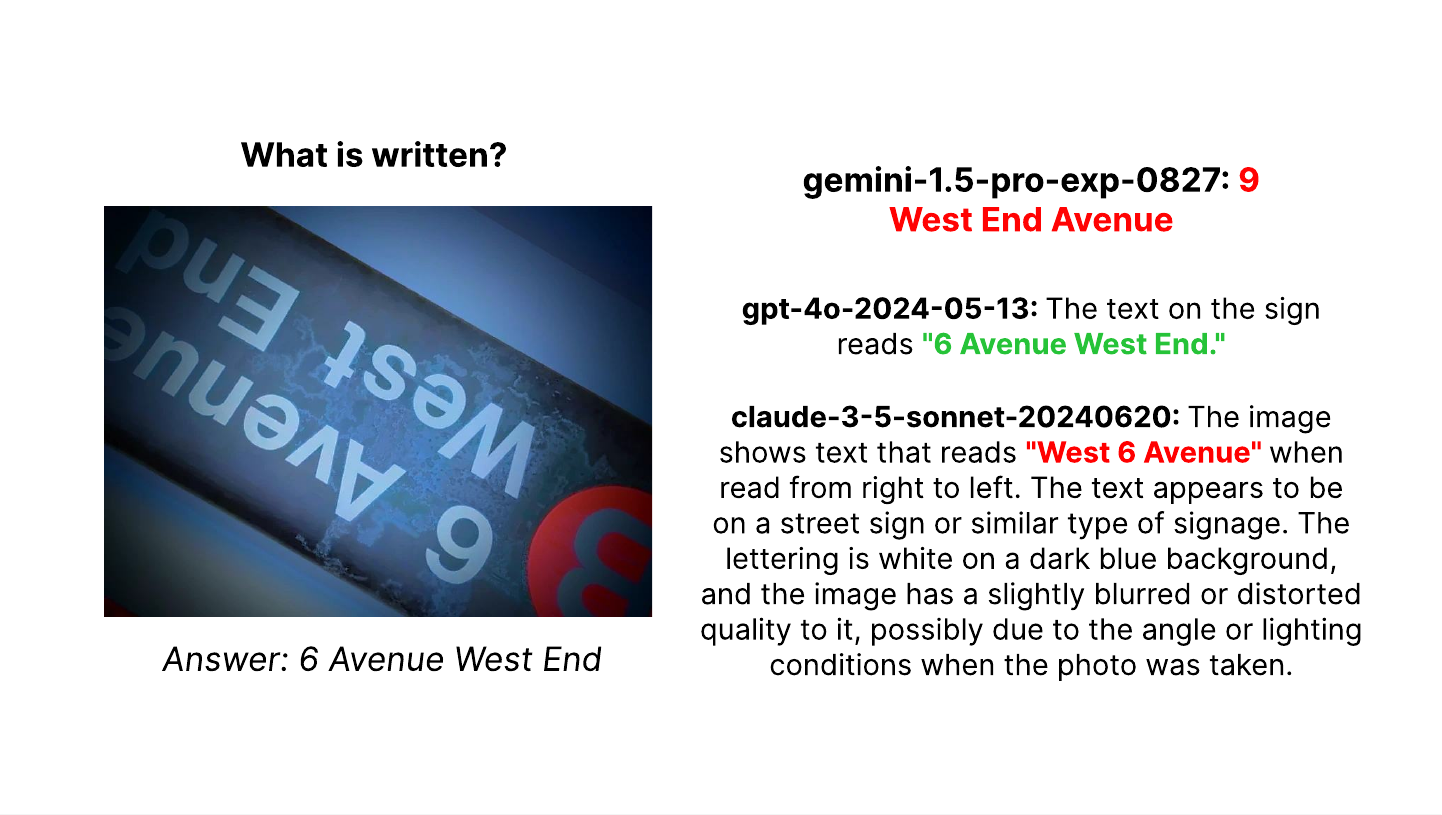}
        \caption{Sign Failure Case}
        \label{fig:sign}
    \end{subfigure}
    \caption{\textbf{Hard OCR Failure Cases}. These failure cases show that models still fail in cases where the text is perturbed such as rotations or messy hand-writing.}
    \label{fig:failure_cases_ocr}
\end{figure*}

\textit{Hard OCR} (\autoref{fig:failure_cases_ocr}). While VLMs perform well at transcribing easily legible text, they struggle with perturbed text (e.g., rotations, blur). Reading such difficult text is essential for real-world applications. We show two failure cases: one with unclear handwriting and another with rotated text.

\begin{figure*}[h]
    \centering
    \begin{subfigure}[b]{\linewidth}
        \centering
        \includegraphics[width=\linewidth]{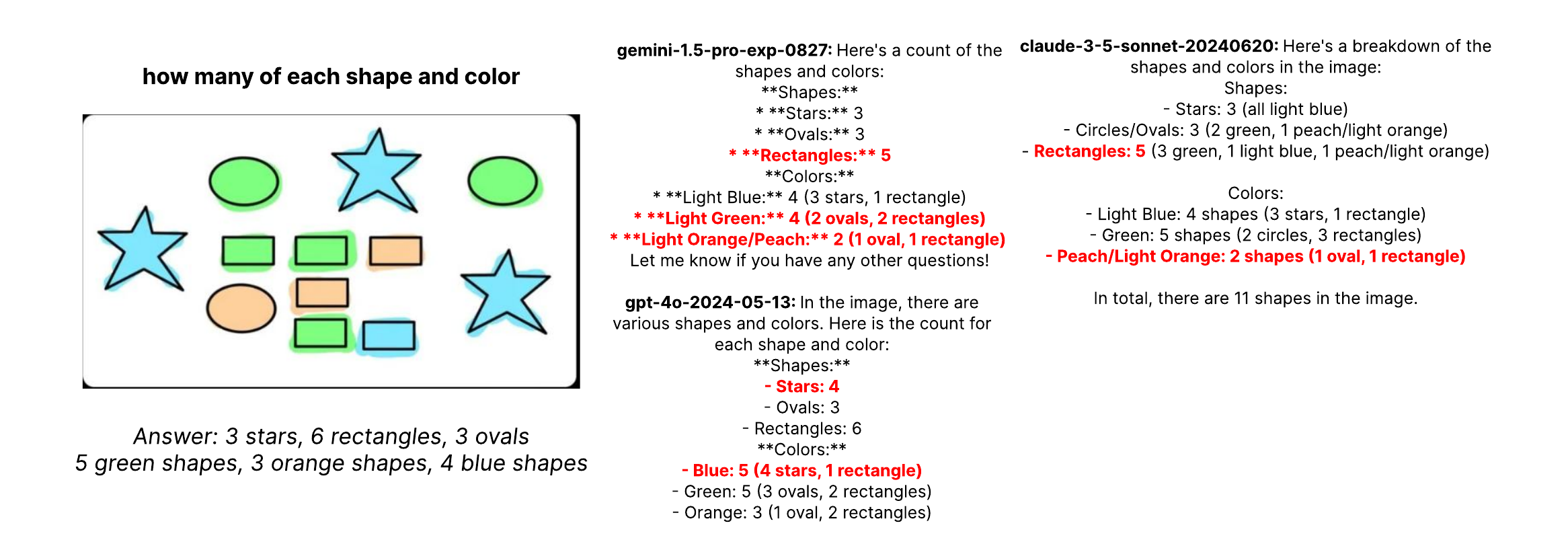}
        \caption{Shapes Failure Case}
        \label{fig:shapes}
    \end{subfigure}
    \hfill
    \begin{subfigure}[b]{\linewidth}
        \centering
        \includegraphics[width=\linewidth]{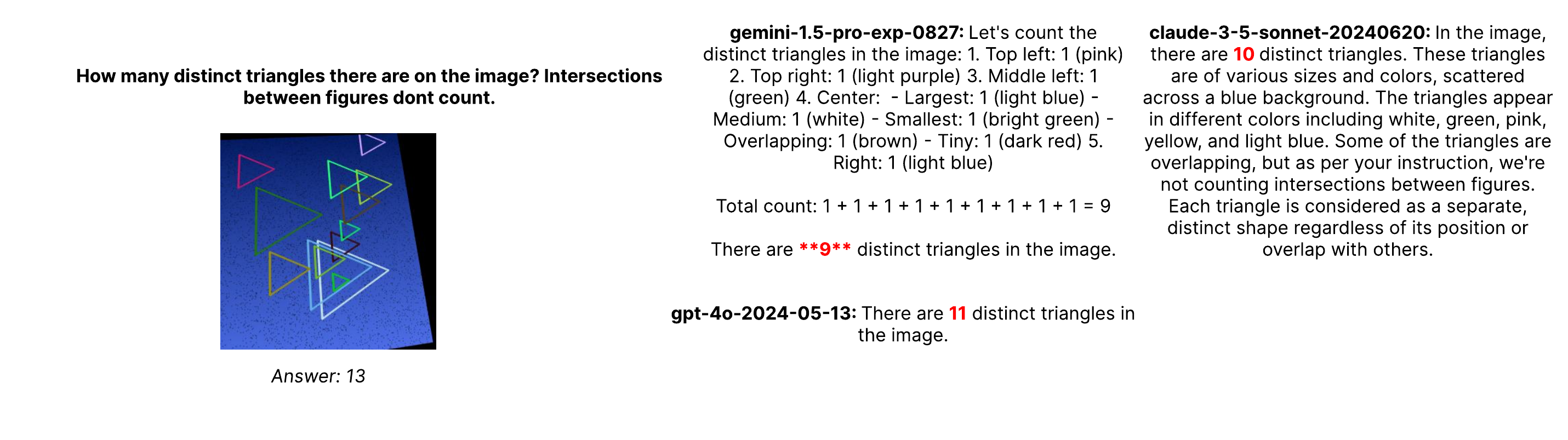}
        \caption{Triangles Failure Case}
        \label{fig:Triangles}
    \end{subfigure}
    \caption{\textbf{Hard Counting Failure Cases}. These cases show that proprietary models still fail at counting tasks involving shapes even when the format is clear.}
    \label{fig:failure_cases_counting}
\end{figure*}

\noindent\textit{Counting} (\autoref{fig:failure_cases_counting}). Counting is a critical skill for decision-making across education, organization, and daily life. While humans count effortlessly, VLMs still struggle. We provide two examples where the top three models fail: one requiring counting based on shape and color, and another involving intersecting triangles.

\begin{figure*}[htbp]
    \centering
    \begin{subfigure}{\linewidth}
        \centering
        \includegraphics[width=\linewidth, height=6cm]{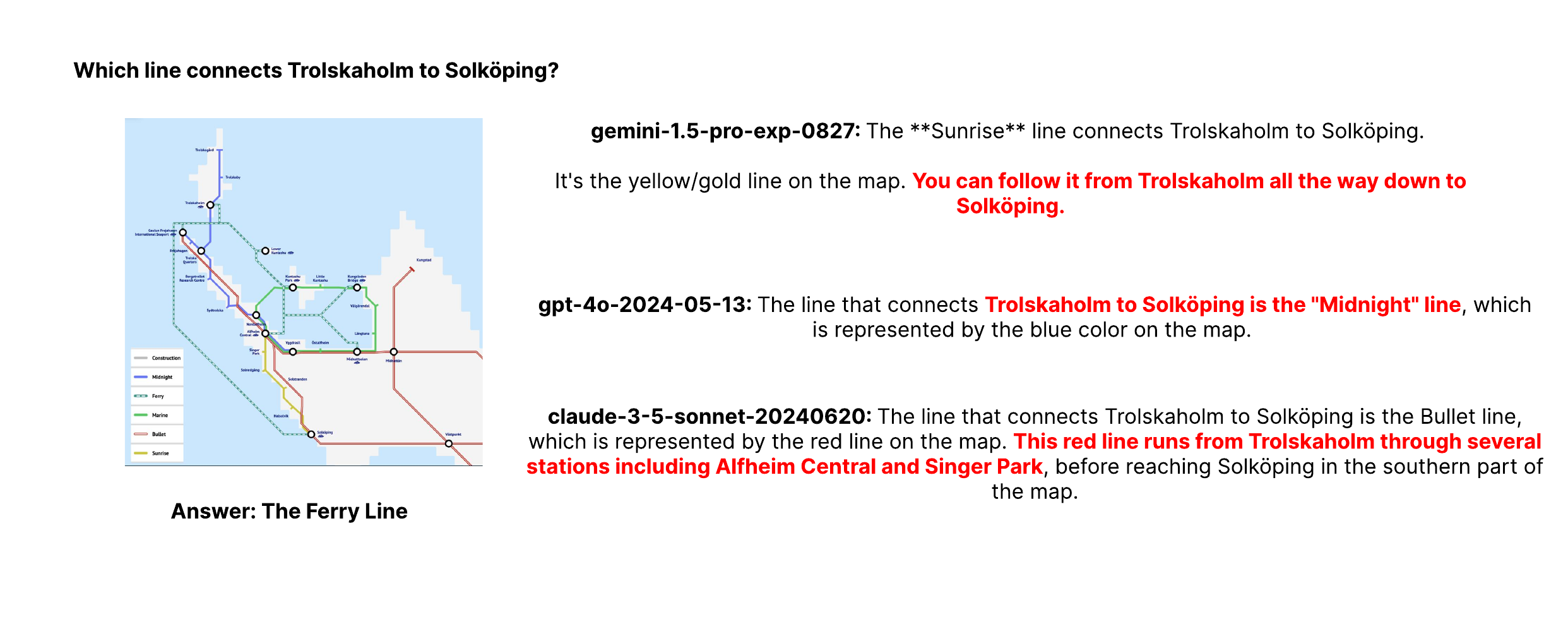}
        \caption{Map Failure Case}
        \label{fig:map}
    \end{subfigure}
    \hfill

    \begin{subfigure}{\linewidth}
        \centering
        \includegraphics[width=\linewidth]{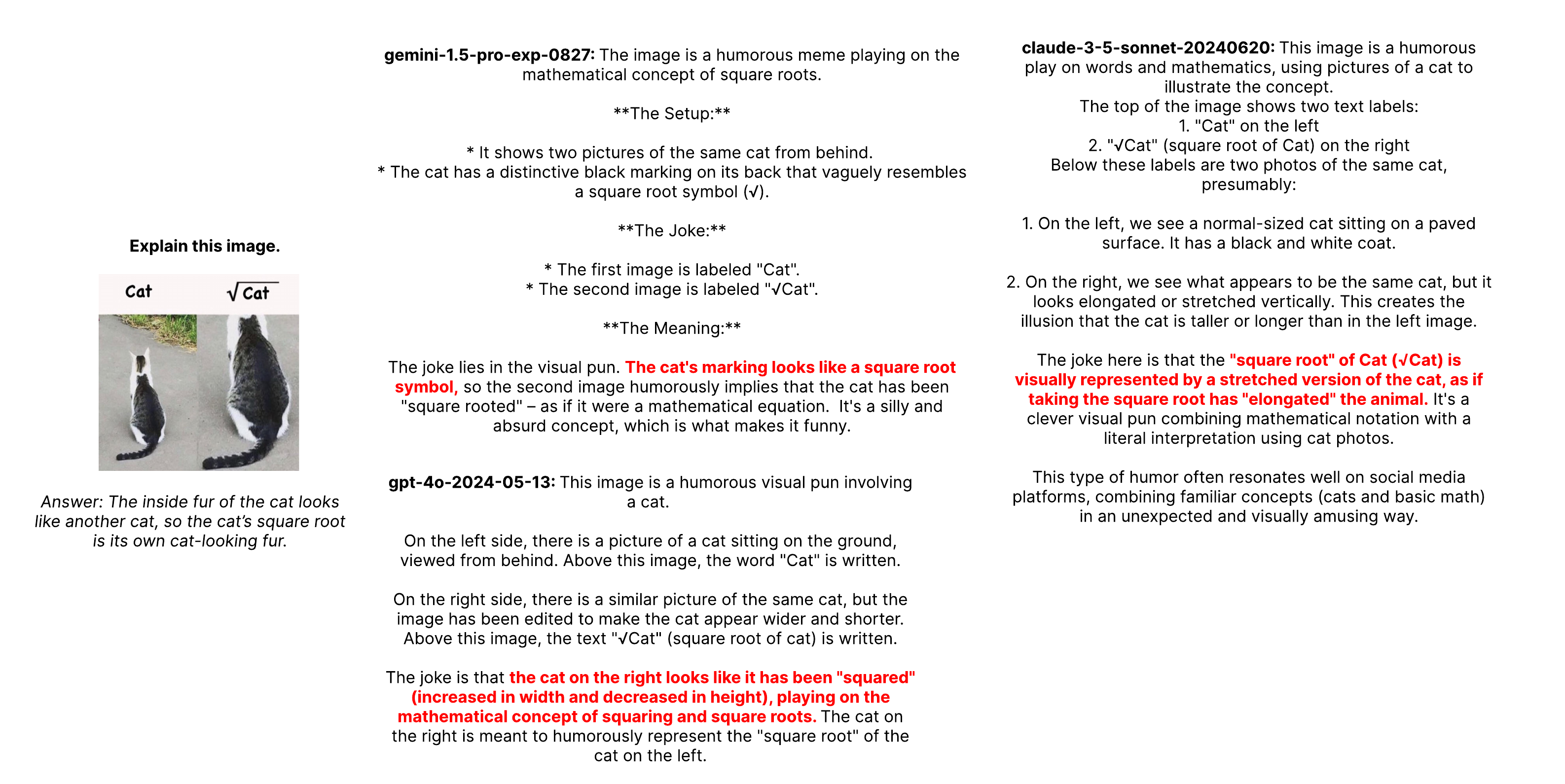}
        \caption{Meme Failure Case}
        \label{fig:meme}
    \end{subfigure}
    \hfill

    \begin{subfigure}{\linewidth}
        \centering
        \includegraphics[width=\linewidth]{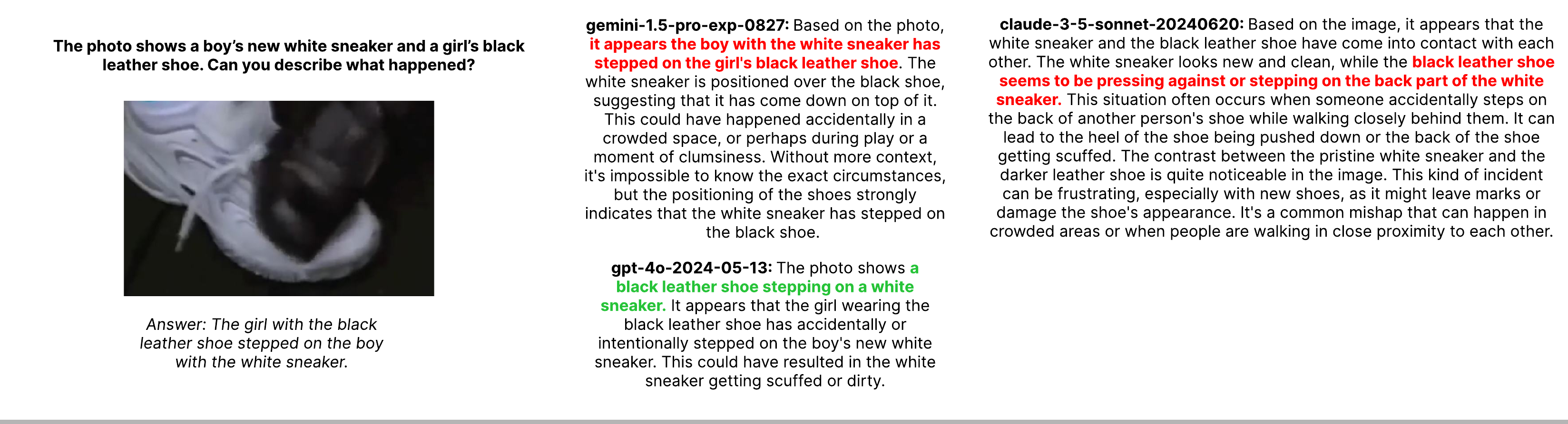}
        \caption{Shoes Failure Case}
        \label{fig:shoes}
    \end{subfigure}

    \caption{\textbf{Hard Reasoning Failure Cases (part 1).} The failure cases highlight the model's inability to connect the visual reasoning with language reasoning. While the vision model identifies an object, it is not able to identify the relationship between them correctly.}
    \label{fig:hard_reasoning_failures}
\end{figure*}

\begin{figure*}
    \begin{subfigure}{\linewidth}
        \centering
        \includegraphics[width=\linewidth]{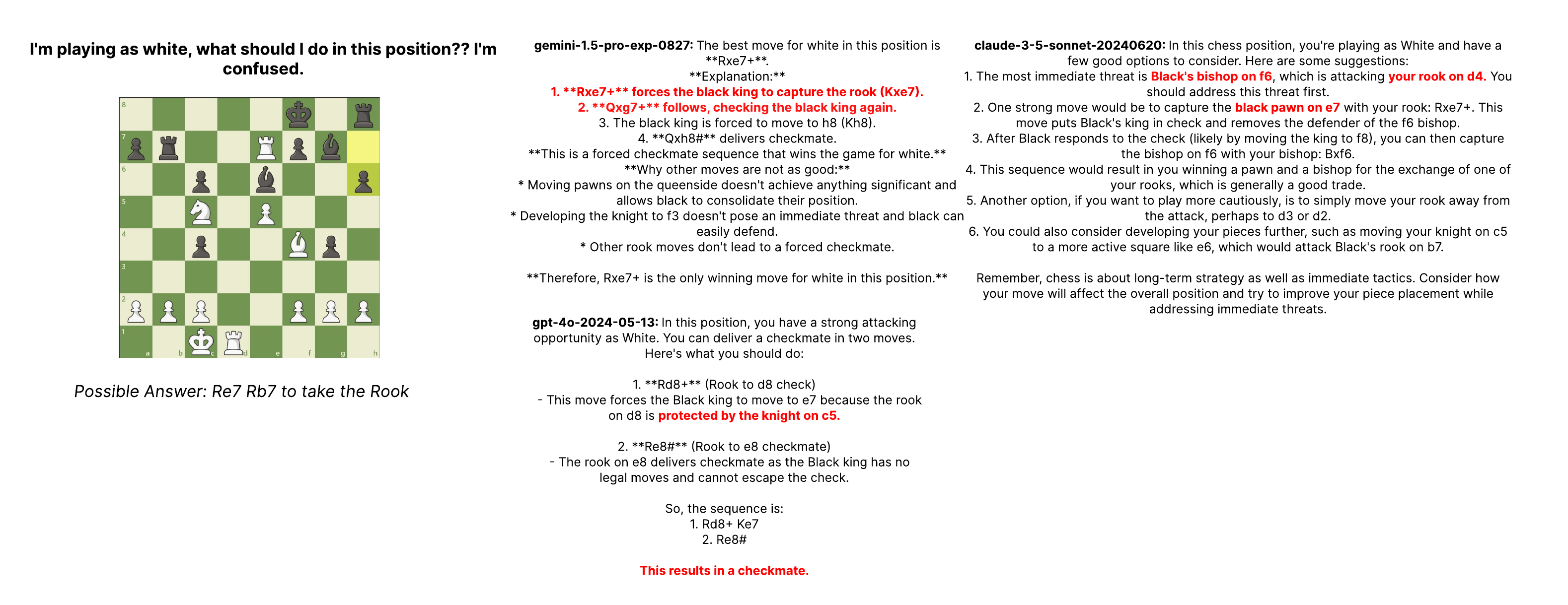}
        \caption{Chess Failure Case}
        \label{fig:chess}
    \end{subfigure}
    
    \begin{subfigure}{\linewidth}
        \centering
        \includegraphics[width=\linewidth]{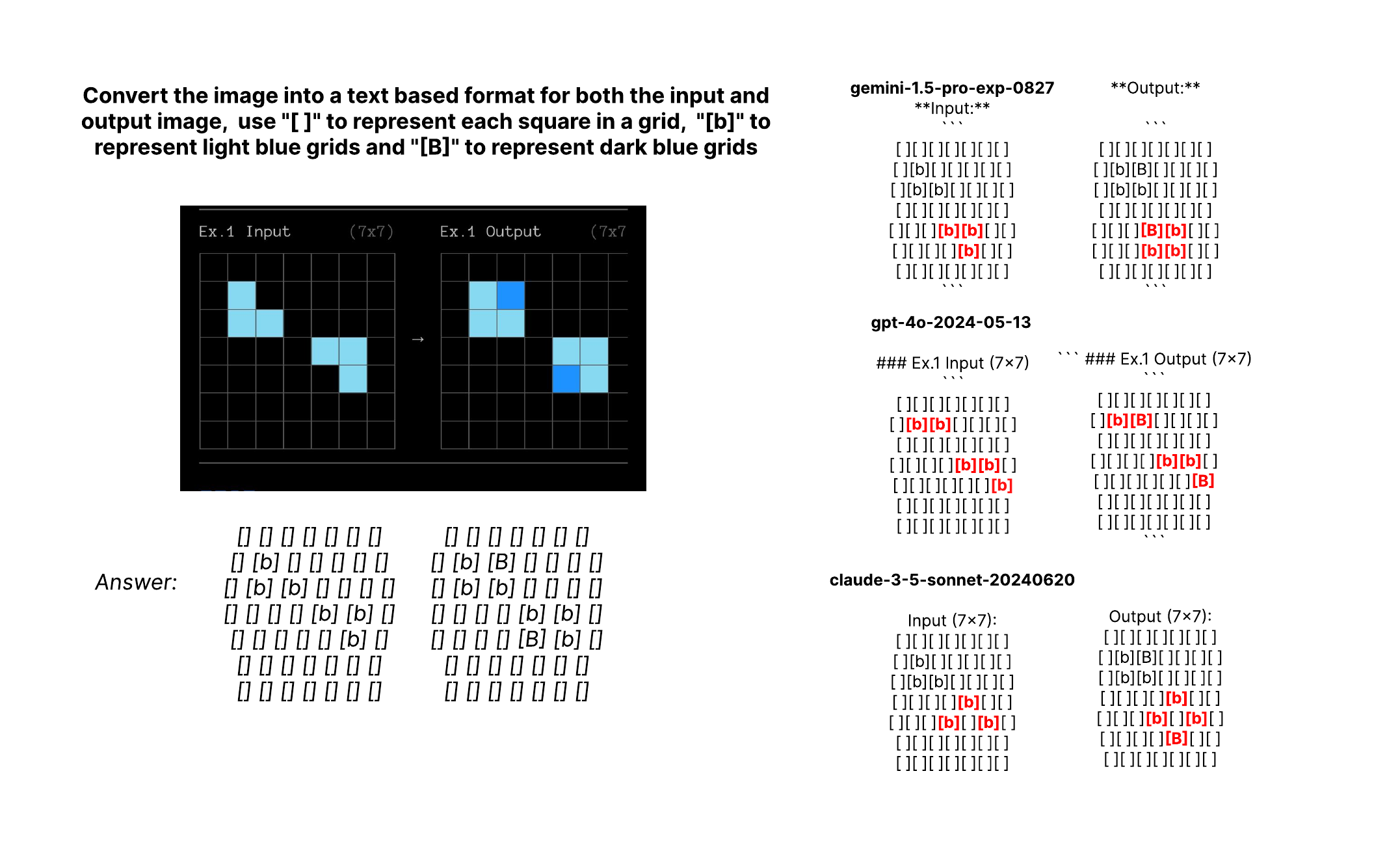}
        \caption{ARC Failure Case}
        \label{fig:arc}
    \end{subfigure}
    \caption{\textbf{Hard Reasoning Failure Cases (part 2).} These failure cases highlight the inability for the model to be able to correctly map out a grid-like structure and the various pieces in it.}
    \label{fig:failure_case_reasoning}
\end{figure*}

\noindent\textit{Reasoning} (\autoref{fig:hard_reasoning_failures}, \autoref{fig:failure_case_reasoning}). Reasoning is essential for helping users tackle complex problems. While reasoning remains a challenge for both VLMs and LLMs, we present five unique VLM failure cases.

\end{document}